\documentclass[sigconf, nonacm]{acmart}

\newcommand\vldbdoi{XX.XX/XXX.XX}
\newcommand\vldbpages{XXX-XXX}
\newcommand\vldbvolume{18}
\newcommand\vldbissue{1}
\newcommand\vldbyear{2025}
\newcommand\vldbauthors{\authors}
\newcommand\vldbtitle{\shorttitle} 
\newcommand\vldbavailabilityurl{https://github.com/nikolaimerkel/reordering}
\newcommand\vldbpagestyle{plain} 

\usepackage{balance}
\usepackage{caption}
\usepackage{subcaption}
\usepackage{pifont}
\usepackage{float}
\usepackage{placeins}
\newcommand{\blue}[1]{\textcolor{black}{#1}}

\newcommand{\dgl}{\emph{DGL}}
\newcommand{\pyg}{\emph{PyG}}
\usepackage{xcolor}

\newcommand{\revision}{\color{black}}
\newcommand{\orginal}{\color{black}}
\usepackage{makecell}

\newcommand{\prof}{\emph{average gap profile}}
\newcommand{\band}{\emph{graph bandwidth}}
\newcommand{\avgb}{\emph{average graph bandwidth}}

\newcommand{\twit}{\emph{twitter}}
\newcommand{\pape}{\emph{papers100}}
\newcommand{\dima}{\emph{dimacs9-USA}}
\newcommand{\live}{\emph{livejournal}}
\newcommand{\orku}{\emph{orkut}}
\newcommand{\redd}{\emph{reddit}}
\newcommand{\socp}{\emph{soc-pokec}}
\newcommand{\produ}{\emph{products}}
\newcommand{\webb}{\emph{web-BerkStan}}
\newcommand{\wiki}{\emph{wikipedia}}

\newcommand{\minla}{\emph{minla}}

\newcommand{\degs}{\emph{degsort}}
\newcommand{\hubs}{\emph{hubsort}}
\newcommand{\slas}{\emph{slashburn}}
\newcommand{\hubc}{\emph{hubcluster}}
\newcommand{\ldg}{\emph{ldg}}
\newcommand{\rcm}{\emph{rcm}}
\newcommand{\dfs}{\emph{dfs}}
\newcommand{\bfs}{\emph{bfs}}
\newcommand{\gord}{\emph{gorder}}
\newcommand{\meti}{\emph{Metis}}
\newcommand{\metii}{\emph{metis-16}}
\newcommand{\metiii}{\emph{metis-128}}
\newcommand{\metiiii}{\emph{metis-1024}}
\newcommand{\metiiiii}{\emph{metis-8192}}
\newcommand{\metiiiiii}{\emph{metis-65536}}
\newcommand{\rabb}{\emph{rabbit}}
\begin{document}
%\title{Can Graph Reordering Speed Up Graph Neural Network Training? An Experimental Study [Experiment, Analysis \& Benchmark]}
\title{Can Graph Reordering Speed Up Graph Neural Network Training? An Experimental Study}

%%
%% The "author" command and its associated commands are used to define the authors and their affiliations.
%\author{Nikolai Merkel}
%\affiliation{%
%  \institution{Technical University of Munich}
%  \streetaddress{}
%  \city{}
%  \state{}
%  \postcode{}
%}
%\email{nikolai.merkel@tum.de}
%
%\author{Pierre Toussing}
%%\orcid{0000-0002-1825-0097}
%\affiliation{%
%  \institution{Technical University of Munich}
%  \streetaddress{}
%  \city{}
%  \state{}
%  \postcode{}
%}
%  \email{pierre.toussing@tum.de}
%
%\author{Ruben Mayer}
%%\orcid{0000-0001-5109-3700}
%\affiliation{%
%  \institution{University of Bayreuth}
% % \city{Rocquencourt}
% % \country{France}
%}
%\email{ruben.mayer@uni-bayreuth.de}
%
%\author{Hans-Arno Jacobsen}
%\affiliation{%
%  \institution{University of Toronto}
%  %\city{T\"ubingen}
%  %\country{Germany}
%}
%\email{jacobsen@eecg.toronto.edu}

%\author{Nikolai Merkel, Pierre Toussing}
%\affiliation{%
%  \institution{Technical University of Munich}
%  \city{}
%  \country{}
%}
%\email{{nikolai.merkel, pierre.toussing}@tum.de}

\author{Nikolai Merkel}
\affiliation{%
  \institution{Technical University of Munich}
  \city{}
  \country{}
}
\email{nikolai.merkel@tum.de}

\author{Pierre Toussing}
\affiliation{%
  \institution{Technical University of Munich}
  \city{}
  \country{}
}
\email{pierre.toussing@tum.de}

\author{Ruben Mayer}
\affiliation{%
  \institution{University of Bayreuth}
  \city{}
  \country{}
}
\email{ruben.mayer@uni-bayreuth.de}

\author{Hans-Arno Jacobsen}
\affiliation{%
  \institution{University of Toronto}
  \city{}
  \country{}
}
\email{jacobsen@eecg.toronto.edu}

%%
%% The abstract is a short summary of the work to be presented in the
%% article.
\begin{abstract}
Graph neural networks (GNNs) are a type of neural network capable of learning on graph-structured data.
However, training GNNs on large-scale graphs is challenging due to iterative aggregations of high-dimensional features from neighboring vertices within sparse graph structures combined with neural network operations.
The sparsity of graphs frequently results in suboptimal memory access patterns and longer training time.
\textit{Graph reordering} is an optimization strategy aiming to improve the graph data layout. 
It has shown to be effective to speed up graph analytics workloads, but its effect on the performance of GNN training has not been investigated yet.
The generalization of reordering to GNN performance is nontrivial, as multiple aspects must be considered: GNN hyper-parameters such as the number of layers, the number of hidden dimensions, and the feature size used in the GNN model, neural network operations, large intermediate vertex states, and GPU acceleration.  

In our work, we close this gap by performing an empirical evaluation of 12 reordering strategies in two state-of-the-art GNN systems, PyTorch Geometric and Deep Graph Library.
Our results show that graph reordering is effective in reducing training time for CPU- and GPU-based training, respectively. 
Further, we find that GNN hyper-parameters influence the effectiveness of reordering, that reordering metrics play an important role in selecting a reordering strategy, that lightweight reordering performs better for GPU-based than for CPU-based training, and that invested reordering time can in many cases~be~amortized. 
\end{abstract}

\maketitle

%%% do not modify the following VLDB block %%
%%% VLDB block start %%%
\pagestyle{\vldbpagestyle}
\begingroup\small\noindent\raggedright\textbf{PVLDB Reference Format:}\\
\vldbauthors. \vldbtitle. PVLDB, \vldbvolume(\vldbissue): \vldbpages, \vldbyear.\\
\href{https://doi.org/\vldbdoi}{doi:\vldbdoi}
\endgroup
\begingroup
\renewcommand\thefootnote{}\footnote{\noindent
This work is licensed under the Creative Commons BY-NC-ND 4.0 International License. Visit \url{https://creativecommons.org/licenses/by-nc-nd/4.0/} to view a copy of this license. For any use beyond those covered by this license, obtain permission by emailing \href{mailto:info@vldb.org}{info@vldb.org}. Copyright is held by the owner/author(s). Publication rights licensed to the VLDB Endowment. \\
\raggedright Proceedings of the VLDB Endowment, Vol. \vldbvolume, No. \vldbissue\ %
ISSN 2150-8097. \\
\href{https://doi.org/\vldbdoi}{doi:\vldbdoi} \\
}\addtocounter{footnote}{-1}\endgroup
%%% VLDB block end %%%

%%% do not modify the following VLDB block %%
%%% VLDB block start %%%
\ifdefempty{\vldbavailabilityurl}{}{
\vspace{.3cm}
\begingroup\small\noindent\raggedright\textbf{PVLDB Artifact Availability:}\\
The source code, data, and/or other artifacts have been made available at \url{\vldbavailabilityurl}.
\endgroup
}
%%% VLDB block end %%%

\section{Introduction}
\label{sec:introdcution}
Graph neural networks (GNNs) have recently emerged as a promising deep learning-based approach to learn on graph-structured data. 
GNNs show great success in different domains such as knowledge graphs~\cite{knowledgegraphs}, recommender systems~\cite{recommendersystems}, and natural language processing~\cite{nlp}.
In GNN training, vertex representations are iteratively learned by aggregating features of the vertices' immediate neighbors followed by neural network transformations.
Through this process, vertex representations are iteratively refined and capture both vertex and graph structure information which can be used for down-stream tasks.
 
Despite the great success of GNNs, the training process is resource-intensive. 
GNNs operate not only on the graph structure itself but also on high-dimensional feature vectors attached to vertices. 
Large intermediate representations are computed for the vertices in each layer leading to large memory overheads. 
Further, neural network operations are performed for each vertex; 
Therefore, GNNs are often trained on GPUs. 
Different GNN hyper-parameters such as the number of layers, the number of hidden dimensions, or the feature size influence the training time.  
Therefore, GNN training differs from traditional graph processing applications such as Breadth-first-search, Connected Components, Shortest Paths, and K-cores which perform lightweight computations, are typically short-running, only lead to small vertex states, are performed on the graph structure only, often exhibit dynamic computational patterns, and are commonly executed on CPUs. 

\textit{Graph reordering} is a technique to optimize the graph's data layout in a way that vertices that are frequently accessed together are stored close to each other in memory to improve locality.
It has been shown that traditional graph processing can benefit from graph reordering \cite{rabbit}. 
\blue{Given that GNN training is often performed on scarce and expensive GPUs, reducing training time through graph reordering can lead to cost savings.}

\blue{Recently, DGI~\cite{DGI}, a framework for easy and efficient GNN \textit{model evaluation} has been proposed which also benefits from graph reordering.}
However, it is not known if graph reordering is effective for GNN training.
It is unclear how GNN-specific parameters and GPU-accelerated training influence the effectiveness of graph reordering. 
Further, it has not been investigated yet if graph reordering quality metrics can be used for graph reordering strategy selection, and if invested graph reordering time can be amortized.
To close this research gap, we perform an extensive experimental study to investigate how effective graph reordering is in speeding up GNN training on CPUs and GPUs.
The main contributions of our work are:
       
\textbf{(1)} We perform extensive evaluations with 12 graph reordering strategies, two predominant GNN systems, different GNN models and GNN hyper-parameters, and 10 real-world graphs.
In our experiments, we perform both GPU-based and CPU-based training. 
Based on our experiments, we find that graph reordering is an effective optimization to reduce GNN training time.
\blue{A high-quality reordering strategy such as \textit{Rabbit} results in speedups of up to 2.19x~(average 1.25x) for CPU-based training and up to 2.43x~(average 1.33x) for GPU-based training. We also investigate the effectiveness of graph reordering when sampling is applied. \textit{Rabbit} achieves speedups of up to 3.68x (average 1.62x) for CPU-based training and up to 3.22x~(average 1.57x) for GPU-based training.}

\textbf{(2)} Graph reordering is a pre-processing task that is performed prior to GNN training. 
In our experiments, we investigate if invested graph reordering time can be amortized. 
We show that for CPU-based training, graph reordering time can be amortized by faster training. 
For GPU-based training, we find that costs can be saved due to reduced training time on monetary expensive GPUs, as the graph reordering can be performed on cheaper CPU machines.

\textbf{(3)} We study the relationship between graph reordering quality metrics and training speedup.
We find that the metrics \prof{} and \avgb{} can be used in many cases to select a graph reordering that leads to a large speedup. 
However, we also find that graph reordering quality metrics are no perfect predictors for GNN speedup indicating that new metrics may be needed to better formalize a graph reordering goal.
    
Our paper is organized as follows.
First, we introduce graph neural network training and graph reordering in Section~\ref{sec:background}. 
In Section~\ref{sec:experimental-methodology}, we describe the methodology of our experimental comparison. 
Then, we analyze our results for both systems in Section~\ref{sec:training-performance} and summarize our main findings in Section~\ref{sec:lessons-learnt}. 
We discuss related work in Section~\ref{sec:related-work}. 
Finally, we conclude our paper in Section~\ref{sec:conclusions}. 

\section{Background}
\label{sec:background}

Let $G=(V, E)$ be a directed graph consisting of a set of vertices $V$ and a set of edges $E=V \times V$. 
The out-neighbors of vertex $v$ are defined as $n^+(v) = \{u | (v,u) \in E \}$, the in-neighbors as $n^-(v) = \{u | (u,v) \in E \}$, and all neighbors of $v$ as $n(v) = n^+(v) \cup n^-(v)$. 
The in- and out-degrees of vertex $v$ are defined as $deg^+(v) = |n^+(v)|$ and $deg^-(v) = |n^-(v)|$, respectively.
The mean degree $deg(G)$ of $G$ is defined as $2 \cdot \frac{|E|}{|V|}$.
%Let $n(v)$ be the neighbors which are connected to $v$.

\subsection{Graph Neural Network Training}
\label{sec:background:gnn}
Graph Neural Networks (GNNs) are a specialized category of neural networks tailored to graph-structured data, leveraging the connections inherent in such data.
In GNN training, vertex representations are iteratively learned as follows:
Initially, each vertex $v$ is characterized by its feature vector $h^{(0)}_v$. 
Then, in each subsequent GNN layer, for each vertex $v$, the learned representations of its neighbors $n(v)$ are aggregated to $a_v^{(k)} = \mathit{AGG}^{(k)}(\{h_u^{(k-1)} | u \in N(v)\})$ and the vertex representation is updated to $h_v^{(k)} = \mathit{UP}^{(k)}(a_v^{(k)},h_v^{(k-1)} )$ by applying an update function $\mathit{UP}$ based on the aggregation $a_v^{(k)}$ and its previous representation $h_v^{(k-1)}$ in layer $k-1$.
This process is \textit{permutation invariant}, meaning the outcome of the learning does not depend on the order of the nodes in the input graph.

\subsection{Graph Reordering}
\label{sec:background:reordering}
\begin{figure}[t]
\centering
\includegraphics[width=\columnwidth]{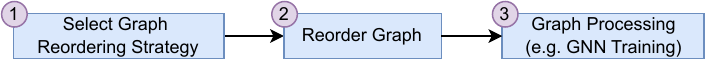}
\vspace{-7mm}
\caption{\blue{Graph Processing Pipeline.}}
\label{fig:pipeline}
\end{figure}

\begin{figure}[!t]
\centering
\subfloat[High locality.]{\includegraphics[width=0.46\columnwidth]{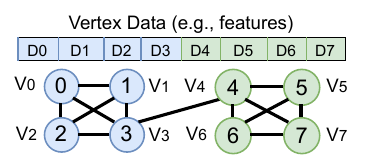}
\label{back:fig:reordering:good-reordering}}
\hfil
\subfloat[Low locality.]{\includegraphics[width=0.46\columnwidth]{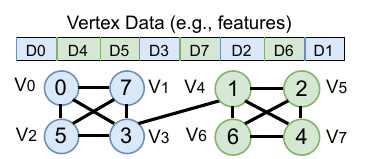}
\label{back:fig:reordering:bad-reordering}}
\vspace{-3mm}
\caption{Two different orderings of the same graph.}
\vspace{-3mm}
\label{back:fig:reordering}
\end{figure}

\blue{Graph reordering is a pre-processing step in a graph processing pipeline (see. Figure~\ref{fig:pipeline})}.
Graph reordering improves memory access locality in graph processing systems by locating vertices that are accessed frequently together close to each other in memory \cite{rabbit}. 
In graph data structures such as compressed sparse row or adjacency matrices, the ID of a vertex is used as a key, e.g., into an array. 
Therefore, the ID of a vertex influences its position in memory, and vertices with close IDs (small difference between the IDs) are also stored close to each other in memory.

The goal of graph reordering is to relabel the vertices of a graph in a way that vertices that are accessed together get close IDs and therefore are stored in consecutive memory. 
This optimization only changes the data layout but does not influence the graph structure or the processing result.
For example, consider the graphs in Figure~\ref{back:fig:reordering}.
The figure shows two different orderings of the \textit{same} graph.
The numbers in the cycles represent the ID of the vertices. 
For example, vertex $v_1$ has ID 1 in the left graph (see Figure~\ref{back:fig:reordering:good-reordering}) and ID 7 in the right graph (see Figure~\ref{back:fig:reordering:bad-reordering}).
The ID of vertex $v_i$ defines the position of its data $D_i$ in the vertex data array (represented above the graph) in which feature vectors or hidden representations of vertices are stored. 
The graph has two dense clusters $c_1$ (colored blue) and $c_2$ (colored green). 
Vertices in a cluster share many neighbors while there is only one edge between the clusters.
In the left ordering of the graph (see Figure~\ref{back:fig:reordering:good-reordering}), vertices of the same cluster have consecutive IDs, and their data is therefore consecutively stored in memory. 
In contrast, in the right ordering of the graph (see Figure~\ref{back:fig:reordering:bad-reordering}), the vertices of a cluster are not labeled consecutively, therefore, their data is spread across the vertex data array.
In GNN training, for vertex $v_0$ the states $D_1$, $D_2$ and $D_3$ of its neighbors $n(v_0) = \{ v_1,v_2,v_3\}$ need to be aggregated, which are in consecutive memory in the left ordering (see blue colored vertex data in Figure~\ref{back:fig:reordering:good-reordering}) but spread across memory in the right ordering (see blue colored vertex data in Figure~\ref{back:fig:reordering:bad-reordering}).
Hence, aggregating the neighbors' data in the first ordering has higher access locality than in the second ordering.

\subsubsection{Graph Reordering Metrics}
\label{sec:background:reordering:metrics}
Different graph reordering metrics exist that describe reordering quality \cite{vertexIISWC}. 
Let $\pi : V \rightarrow V$ be a bijection mapping vertices to their IDs. 
The \textit{gap} between two vertices $u, v \in V$ is defined as $\xi_{\pi}(u, v) = |\pi(u) - \pi(v)|$. The larger the gap between two vertices, the more distant they are from each other. 
The \textit{vertex bandwidth} for a vertex $v \in V$ is defined as $\beta_v(G,\pi)=\mathit{max} \{ \xi_{\pi}(v, u) | \forall u \in n(v)\}$, meaning it is defined by the ID of $v$ and the ID of its most distant neighbor. 
In the following, three quality metrics (lower values are better) are defined:
The \textit{average gap profile} for graph $G$ is defined as $\xi(G, \pi) = \frac{1}{|E|} \sum_{u,v \in E} \xi_{\pi}(u, v)$.
The \textit{graph bandwidth} of graph $G$ is defined as:$\beta(G,\pi) = \mathit{max}\{ \xi_{\pi}(v, u) | \forall (v,u) \in E \}$.
The \textit{average graph bandwidth} of graph $G$ is defined as: $\widehat{\beta}(G, \pi) = \frac{1}{|V|}\sum_{v \in V}\beta_v(G,\pi)$.

\subsubsection{Graph Reordering Strategies}
\label{sec:background:reordering:strategies}

\begin{table}[htbp]
\caption{Graph reordering strategies used in our study.}
\vspace{-4mm}
\begin{center}
\begin{tabular}{|l|l|}
\hline
\textbf{Reordering Category} & \textbf{Reordering Strategies}  \\
\hline
\hline
GAP-based  & MINLA  \\ \hline
%Degree- and Hub-based   &    \makecell[l]{Degree Sort, Hub Sort, Hub Cluster,\\ \hspace{4em}SlashBurn}  \\ \hline
Degree- and Hub-based   &    \makecell[l]{Degree Sort, Hub Sort, Hub Cluster, \\ SlashBurn}\\ \hline
Window-based &     Gorder   \\ \hline
Partition-based     &  Rabbit, LDG, Metis     \\ \hline
Fill-reducing-based     &  RCM, DFS, \blue{BFS}   \\ \hline
\end{tabular}
\label{tab:background:reorderings}
\end{center}
\end{table}

Graph reordering is a vibrant research field and many different graph reordering strategies exist \cite{gorder, rcm,rabbit,slashburn, experimentsminla, vertexIISWC,zhang}. 
According to Barik et al~\cite{vertexIISWC}, graph reordering strategies can be categorized into the following categories: 
(1)~Degree- and Hub-based graph reordering approaches mainly use the degree information of vertices for reordering. 
(2)~Partition-based approaches divide the vertex set into partitions with the goal of minimizing the number of edges between partitions and balancing the number of vertices per partition. 
This category also includes community detection and graph clustering approaches, that detect densely connected clusters in graphs. 
Then, the reordering is performed based on the partitions or clusters, e.g., vertices of the same partition or cluster get consecutive IDs. 
(3)~Window-based approaches slide a window over the vertices and maximize a score for the window.
(4)~GAP-based approaches minimize the \textit{average gap profile}. 
(5)~Fill-reducing-based approaches reorder a matrix in a way that the number of non-zero elements in the factorized matrix is minimized.

In the following, we introduce for each category graph reordering strategies. 
An overview can be found in Table~\ref{tab:background:reorderings}. 

\textbf{Rabbit:} Rabbit order is a lightweight community detection-based graph reordering strategy. 
Many real-world graphs have a community structure consisting of densely connected vertices that are accessed together in the processing. % that are frequently accessed together. 
Rabbit order exploits this property of real-world graphs for reordering \cite{rabbit}.

\textbf{MINLA:} 
The Minimum Linear Arrangement Problem (MINLA) \cite{experimentsminla} is a combinatorial optimization problem with the goal of minimizing the \avgb{} (as introduced in Section~\ref{sec:background:reordering:metrics}). %$\sum_{u,v \in E} |\pi(u) - \pi(v)|$ \cite{experimentsminla}. %The problem is NP-complete \cite{npcompleteminla}.

\textbf{SlashBurn:} SlashBurn iteratively identifies and removes hubs from the graph to create many small disconnected components. 
The goal is to bring the adjacency matrix to block-diagonal~form~\cite{slashburn}.

\textbf{Gorder:} Gorder \cite{gorder} is window-based graph reordering strategy. 
Let $S_s(u, v) = |n^-(u) \cap n^-(v)|$ be the number of common in-neighbors of vertices $u$ and $v$, and let $S_n(u, v)$ be the number of times $u$ and $v$ are connected by an edge which can be either 0, 1 or 2, and let $w$ be the window size. 
Gorder maximizes the "Gscore" which is defined as $\sum_{0 < \pi(v) - \pi(u) \leq w } (S_n(u, v) + S_s(u, v))$.

\textbf{Degree Sort:} Degree Sort is a lightweight approach where the vertices of the graph are sorted by degree and are consecutively labeled in that order~\cite{when}.

\textbf{Hub Sort:} Similar to Degree Sort, in Hub Sort, vertices with a degree larger than a given threshold (Hub vertices) are sorted by degree and placed next to each other~\cite{zhang}. %while the remaining vertices keep their original label in most cases \cite{zhang}.

\textbf{Hub Clustering:} Similar to Hub Sort with the difference that hub vertices are not sorted but still placed next to each other in their original relative order~\cite{when}. 

\textbf{RCM:} Reverse-Cuthill-McKee (RCM) is a breadth-first-search (BFS) based approach. 
Different from BFS, vertices are sorted by degree before being added to the queue. 
The goal of RCM is to reduce the graph bandwidth~\cite{rcm}. 
 
\textbf{DFS:} The graph is traversed with depth-first search, and the vertices are labeled in the order in which they are visited.

\blue{\textbf{BFS:} The graph is traversed with breadth-first search, and the vertices are labeled in the order in which they are visited.}

\textbf{LDG:} Linear Deterministic Greedy (LDG) \cite{ldg} is a score-based streaming vertex partitioner. 
Vertices are streamed one after the other along with their neighbors and are assigned to partitions on the fly based on a score. 
A vertex $v$ is assigned to partition $i$ for which the highest score $s(i) = |P_i \cap n(v)| \cdot (1-\frac{|P_i|}{C})$ is achieved. 
$P_i$ is the set of vertices already assigned to partition $i$ and $C$ is a capacity limit per partition, e.g., $C=\frac{|V|}{k}$ with $k$ being the number of partitions. 
In the first part of the equation, partitions are preferred to which neighbors of $v$ are already assigned, while the second part ensures balancing so that partitions are of similar size.  

\textbf{Metis:} Metis is an in-memory vertex partitioner. 
It uses a multilevel technique consisting of three phases~\cite{metis}. 
First, the graph is coarsened, and thereby the size of the graph is reduced. 
Then, the coarsened graph is partitioned. 
Finally, the partitioning is projected back to the original graph and refined.

\section{Experimental Methodology}
\label{sec:experimental-methodology}
We want to answer the following research questions:
\begin{description}
    \item[\textbf{Q1}] Is graph reordering effective in reducing GNN training~time?
    \item[\textbf{Q2}] How do GNN parameters such as \textit{number of layers}, \textit{number of hidden dimensions}, and \textit{feature size} influence the effectiveness of graph reordering for GNN training?
    \item[\textbf{Q3}] \blue{How does the effectiveness of graph reordering vary across different hardware configurations, including CPU-based training and GPU architectures?}
    \item[\textbf{Q4}] Can invested graph reordering time be amortized by faster GNN training?
    \item[\textbf{Q5}] Are existing graph reordering quality metrics good indicators for training speedup, and can they be used to select a graph reordering strategy?
    \item[\textbf{Q6}] \blue{Do inherent graph characteristics, such as community structure and density, influence the effectiveness of graph reordering?}
    \item[\textbf{Q7}] \blue{Is graph reordering effective when sampling is applied?}
\end{description}

In the following, we introduce our experimental setup to answer the research questions by describing which GNN systems, graph reordering strategies, GNN models, GNN parameters, infrastructure (GPU or CPU), and datasets we selected for our study, and which metrics we measured. 

\textbf{Systems:} We perform experiments with the predominant GNN systems Deep Graph Library~(DGL)~\cite{dgl} and PyTorch Geometric~(PyG)~\cite{pyg}. 
Both systems are commonly used for GNN training and have large user communities\footnote{\dgl{} and \pyg{} have 12.4k and 18.9k stars on GitHub, respectively. Further, \dgl{} and \pyg{} are forked 2.9k and 3.4k times on GitHub, respectively.}, indicating that optimizing both systems with graph reordering is useful to a large user base. 
Further, both systems provide highly optimized CPU- and \mbox{GPU-based~training}.

\textbf{Reordering Strategies:} We selected all 12 graph reordering strategies from five different categories which were introduced in Section~\ref{sec:background:reordering} (see Table~\ref{tab:background:reorderings} for an overview). 
This selection covers at least one representative of each category.  
For graph reordering with \meti{}, we observed that the number of partitions can influence the effectiveness, therefore, we selected different numbers of partitions $k \in \{16, 128, 1024, 8192, 65536\}$, and treated each configuration as a different reordering \textit{metis-16}, \textit{metis-128}, \textit{metis-1024}, \textit{metis-8192}, and \textit{metis-65536}. 
Therefore, we create in total 16 different reorderings per graph. Our common baseline is random graph reordering.

\textbf{GNN architectures:} We selected two representative GNN architectures, GAT \cite{gat} and GCN \cite{gcn} which are commonly used \cite{dgl, pyg}. 
\revision{}
For both GNN models, the hyperparameters \textit{number of hidden dimensions} and \textit{number of layers} need to be configured. 
Our literature review indicates that the number of hidden dimensions typically ranges between 16 and 256. 
Regarding the number of layers, we find that most GNNs are designed with 2 layers as the default, followed by 3 layers, and only in rare instances, 4 layers \cite{DGI,DeepGalois,P3,ROC,PaGraph,GNNLab,distdgl,dgl}. 
We also observe that the number of hidden dimensions is usually smaller than the feature size. 
Accordingly, we selected the number of hidden dimensions as 16, 32, 64, and 256, and the number of layers as 2, 3, and 4, to encompass the range of commonly used parameters.
\orginal{}

\textbf{Datasets:} We selected graphs from different categories to investigate whether the graph reordering effectiveness depends on the graph type. 
Table~\ref{tab:graphs} reports the graphs along with the number of vertices and edges, the mean degree, and the average local clustering coefficient\footnote{Let $t(v)$ be the number of triangles connected to vertex $v$, and $tr(v) = 0.5 \cdot deg(v) \cdot (deg(v)-1)$ be the number of triplets connected to $v$. According to \cite{newman2003structure}, the local clustering coefficient of vertex $v$ is defined as $c(v) = \frac{t(v)}{tr(v)}$ and the average local clustering of graph G as $C(G) = \frac{1}{|V|} \cdot \sum_{v \in V} c(v)$.}. 
In order to investigate the influence of \textit{the number of features} on the graph reordering effectiveness, we set the feature size for all graphs to three different values 16, 64 and 512.

\begin{table}[t]
\caption{Used graphs along with the number of vertices $|V|$, number of edges $|E|$, mean degree (MD) and average local clustering coefficient (LCC).}
\vspace{-4mm}
\begin{center}
\begin{tabular}{|l|c|c|c|c|}
\hline
\textbf{Graph} &  $\mathbf{|V|}$ &    $\mathbf{|E|}$  &  \textbf{MD} &  \textbf{LCC}\\
\hline
\hline
web-BerkStan \cite{snapnets}&      0.69M &   6.65M &       19.41 &        0.63 \\ \hline
   soc-pokec\cite{snapnets} &      1.63M &  22.30M &       27.32 &        0.12 \\ \hline
 dimacs9-USA \cite{konnekt} &     23.95M &  28.85M &        2.41 &        0.02 \\ \hline
 livejournal\cite{snapnets} &      4.00M &  34.68M &       17.35 &        0.35 \\ \hline
      reddit \cite{hamilton2017inductive} &    0.23M &  57.31M &      491.99 &        0.58 \\ \hline
    products \cite{Bhatia16, hu2020open} &      2.40M &  61.86M &       51.54 &        0.44 \\ \hline
   wikipedia \cite{konnekt} &      3.60M &  77.58M &       43.06 &        0.25 \\ \hline
       orkut \cite{snapnets} &      3.07M & 117.19M &       76.28 &        0.17 \\ \hline
\end{tabular}
\label{tab:graphs}
\end{center}
\end{table}

\textbf{Metrics:} We selected different graph reordering metrics (\prof{}, \band{} and \avgb{}) introduced in Section~\ref{sec:background} to compare graph reordering strategies against each other, to investigate their relationship with GNN training speedup, and to explore their utility for graph reordering selection. 
Further, we measure graph reordering time and \mbox{GNN training time.}

\textbf{Infrastructure:} We use different hardware setups to investigate the effectiveness of graph reordering for CPU- and GPU-based training. 
For GPU-based training, we use a machine with 256~GB main memory, 64 cores, and a Nvidia RTX 8000 GPU with 48~GB of VRAM.  
For CPU-based GNN training, we use a machine with 164 GB main memory and 8 cores. 
This machine is also used \mbox{for graph reordering}.

\section{Results}
\subsection{Overall Performance}
\label{sec:training-performance}

\begin{figure}[t]
\begin{subfigure}[b]{1\linewidth}
\centering
\includegraphics[width=\linewidth]{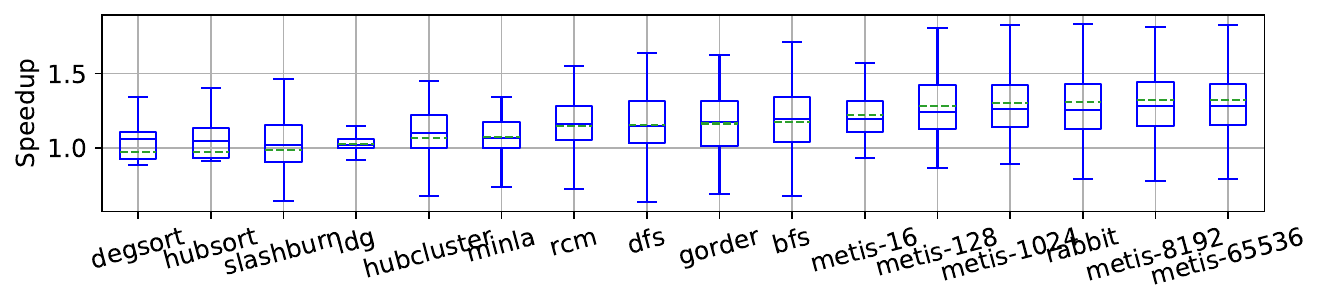}
\vspace{-6mm}
\caption{\blue{Speedup distribution. Mean represented as dashed line.}}
\label{fig:overview:speedups}
\end{subfigure}
\hfill
\begin{subfigure}[b]{1\linewidth}
\centering
\includegraphics[width=\linewidth]{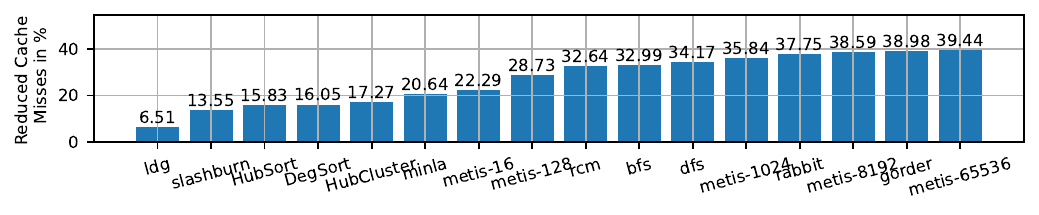}
\vspace{-6mm}
\caption{Reduction of cache misses in percent (CPU).}
\label{fig:overview:cachemisses-cpu}
\end{subfigure}
\hfill
\begin{subfigure}[b]{1\linewidth}
\centering
\includegraphics[width=\linewidth]{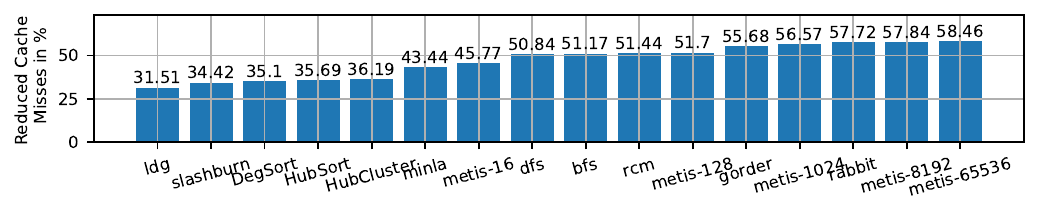}
\vspace{-6mm}
\caption{Reduction of cache misses in percent (GPU).}
\label{fig:overview:cachemisses-gpu}
\end{subfigure}
\caption{Overview of the graph reordering strategies in terms of speedup and cache misses. Larger values are better.}
\label{fig:overview}
\end{figure}

\blue{In the following, we report the average speedups for a 2-layer GNN with a medium feature size of 64. 
The number of hidden dimensions is set to 16 as it is usually smaller than the feature size. 
Subsequently, we will vary the hyper-parameters to investigate the effectiveness of graph reordering under different configurations.}

Figure~\ref{fig:overview:speedups} gives an overview of the average speedup for all graph reordering strategies to indicate which speedups can be expected on average from which strategy.  
We observe that the degree-based approaches \degs{}, \hubs{}, \slas{}, lead to speedups of 0.97x, 0.98x, and 0.99x, meaning on average they slow down the training. 
The partitioning-based approaches \rabb{}, \metii{}, \metiii{}, \metiiii{}, \metiiiii{}, and \metiiiiii{} are effective in reducing GNN training time and lead to average speedups of 1.31x, 1.22x, 1.28x, 1.30x, 1.32x, and 1.33x, respectively, followed by \bfs{}, \gord{}, \dfs{}, \rcm{}, \minla{}, \hubc{}, and \ldg{} with speedups of 1.18x, 1.16x, 1.15x, 1.15x, 1.07x, and 1.03x, respectively.
For \meti{}, we observe that the more partitions are used, the larger the speedup, indicating that the number of partitions is an important parameter that influences the effectiveness.

For two graphs, \produ{} and \webb{}, we measured cache misses both on the CPU and GPU to investigate if the graph reordering strategies that lead to larger speedups also lead to fewer cache misses. 
To count cache misses on the GPU, we identified the kernels that are most influenced by graph reordering and profiled the cache misses of these kernels. 
Both on the CPU and GPU, we observed that better performing graph reordering strategies lead to fewer cache misses, e.g., the best performing graph reordering strategy \metiiiiii{} reduced the number of cache misses by 39.44\% and 58.46\% on the CPU and GPU, \mbox{respectively (see Figures~\ref{fig:overview:cachemisses-cpu} and \ref{fig:overview:cachemisses-gpu}).}

\revision{}
These results are plausible given the nature of GNN training. During GNN training, vertices iteratively aggregate features and hidden representations from neighboring vertices. Hence, vertices that are frequently accessed together, such as those within the same community, should have contiguous IDs to ensure locality in memory. Partitioning-based approaches like \meti{} and \rabb{} are specifically designed to identify communities, assigning similar IDs to vertices within the same community. This results in improved training speed and reduced cache misses.
Other approaches, such as \bfs{}, \dfs{}, or \rcm{}, also yield decent results as they preserve some locality by traversing the graph in a manner that inherently maintains structure. However, degree-based approaches are less effective. For instance, sorting vertices by degree and relabeling them in that order fails to respect the community structure, resulting in a data layout that does not reflect this structure.
In Section~\ref{sec:quality-metrics}, we compare the reordering approaches based on quality metrics that indicate data locality. Our findings show that partitioning-based approaches indeed achieve much better reordering quality metrics compared to degree-based approaches.

We conclude that \textbf{graph reordering is an effective optimization for reducing cache misses on both CPU and GPU, significantly accelerating GNN training. Particularly, the partitioning-based approaches \meti{} and \rabb{} demonstrate substantial potential}.
\orginal{}

\subsection{Influence of GNN Parameters}
In the following, we investigate how the GNN parameters \textit{hidden dimension}, \textit{number of layers}, and \textit{feature size} influence the effectiveness of graph reordering for GNN training.

\subsubsection{Number of hidden dimensions}
\revision{}
In the following experiments, we increase the number of hidden dimensions from 16 to 32, 64, and 256. 
The number of layers is set to 2, and the feature size is fixed at 256, considering that the number of hidden dimensions is typically smaller than the feature size.
\orginal{}

\begin{figure*}[t]
\centering
\begin{subfigure}[b]{\linewidth}
\centering
\includegraphics[width=\linewidth]{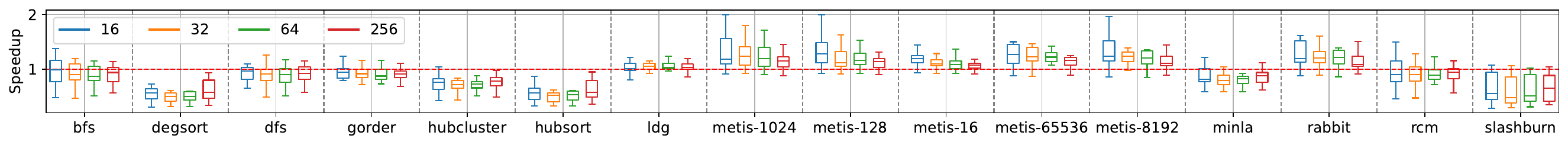}
\vspace{-5mm}
\caption{\blue{CPU-based training.}}
\label{fig:vary-hidden-dimensions:dgl-cpu-gat-gcn}
\end{subfigure}
\hfill
\begin{subfigure}[b]{\linewidth}
\centering
\includegraphics[width=\linewidth]{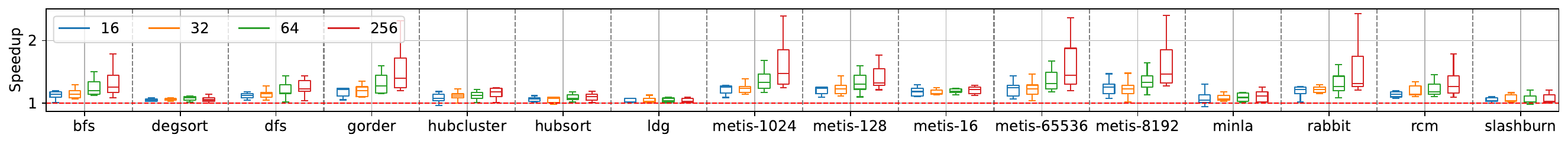}
\vspace{-5mm}
\caption{\blue{GPU-based training (GCN).}}
\label{fig:vary-hidden-dimensions::dgl-gpu-gcn}
\end{subfigure}
\hfill
\begin{subfigure}[b]{\linewidth}
\centering
\includegraphics[width=\linewidth]{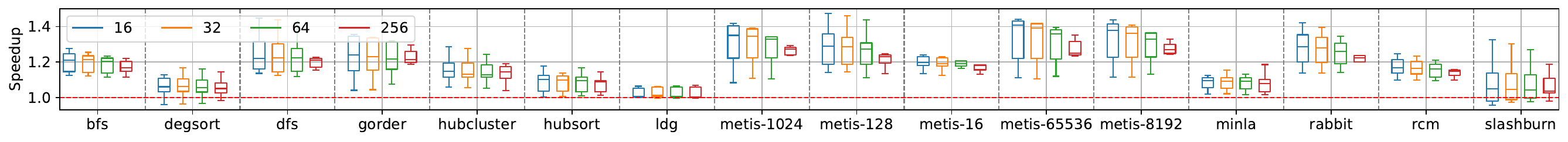}
\vspace{-5mm}
\caption{\blue{GPU-based training (GAT).}}
\label{fig:vary-hidden-dimensions::dgl-gpu-gat}
\end{subfigure}
\vspace{-8mm}
\caption{\blue{Increasing the number of hidden dimensions from 16 to 32, 64, and to 256 in \dgl{}.}}
\label{fig:vary-hidden-dimensions}
\end{figure*}

\begin{figure*}[t]
\centering
\begin{subfigure}[b]{\linewidth}
\centering
\includegraphics[width=\linewidth]{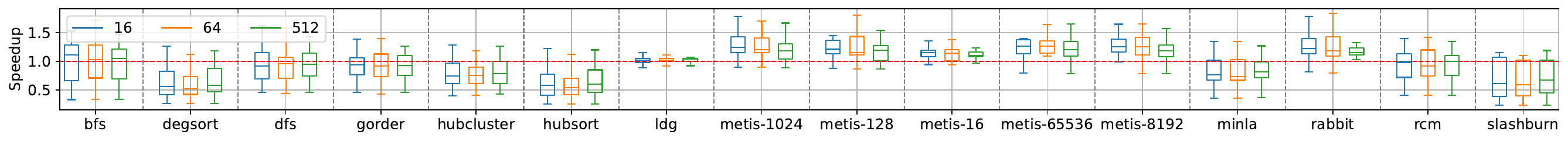}
\vspace{-5mm}
\caption{\blue{CPU-based training.}}
\label{fig:vary-features-dimensions:dgl-cpu-gat-gcn}
\end{subfigure}
\hfill
\begin{subfigure}[b]{\linewidth}
\centering
\includegraphics[width=\linewidth]{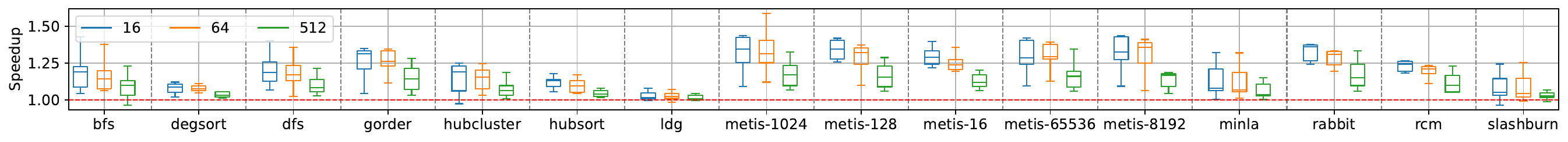}
\vspace{-5mm}
\caption{\blue{GPU-based training (GCN).}}
\label{fig:vary-features-dimensions:dgl-gpu-gcn}
\hfill
\begin{subfigure}[b]{\linewidth}
\centering
\includegraphics[width=\linewidth]{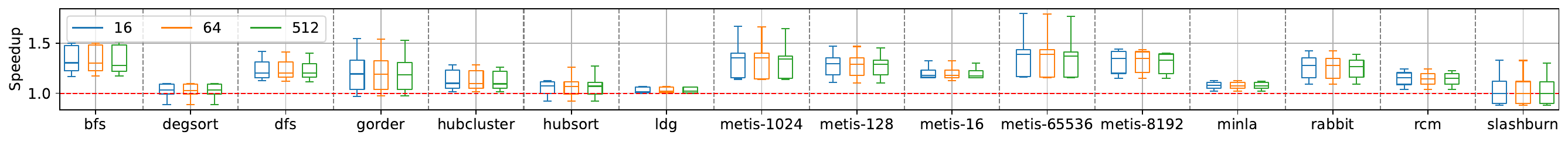}
\vspace{-5mm}
\caption{\blue{GPU-based training (GAT).}}
\label{fig:vary-features-dimensions:dgl-gpu-gat}
\end{subfigure}
\end{subfigure}
\vspace{-8mm}
\caption{\blue{Increasing the feature size from 16 to 64 and to 512 in \dgl{}. Larger values are better.}}
\label{fig:vary-hidden-dimensions}
\end{figure*}
\begin{figure*}[t]
\begin{subfigure}[b]{\linewidth}
\centering
\includegraphics[width=\linewidth]{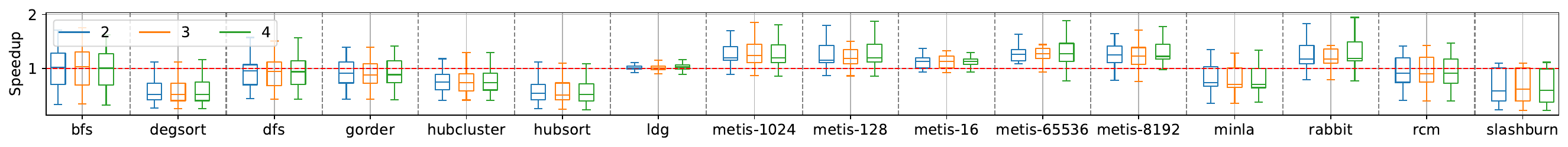}
\vspace{-5mm}
\caption{\blue{CPU-based training.}}
\label{fig:vary-layer-dimensions:dgl-cpu-gat-gcn}
\end{subfigure}
\hfill
\begin{subfigure}[b]{\linewidth}
\centering
\includegraphics[width=\linewidth]{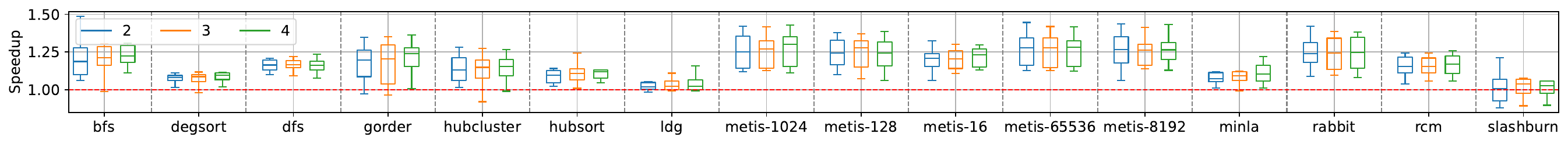}
\vspace{-5mm}
\caption{\blue{GPU-based training.}}
\label{fig:vary-layer-dimensions:dgl-gpu-gat-gcn}
\end{subfigure}
\vspace{-8mm}
\caption{\blue{Increasing the number of layers from 2 to 3 and to 4 in \dgl{}. Larger values are better.}}
\label{fig:vary-hidden-dimensions}
\end{figure*}

\begin{figure}[h]
\centering
\begin{subfigure}[b]{\linewidth}
\centering
\includegraphics[width=\linewidth]{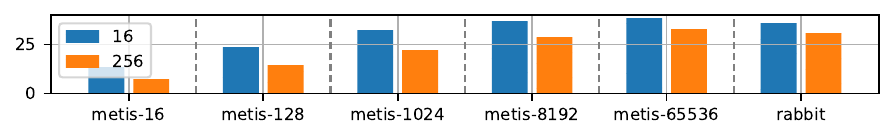}
\vspace{-6mm}
\caption{Increasing the number of hidden dimensions from 16 to 256.}
\label{fig:cachemisses:cpu:hidden}
\end{subfigure}
\begin{subfigure}[b]{\linewidth}
\centering
\includegraphics[width=\linewidth]{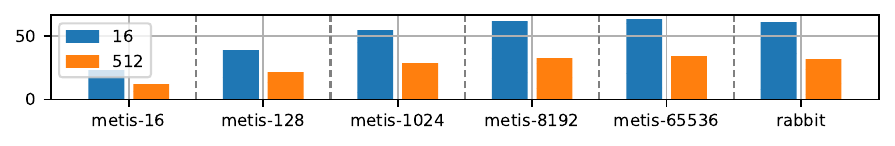}
\vspace{-6mm}
\caption{Increasing the feature size from 16 to 512.}
\label{fig:cachemisses:cpu:features}
\end{subfigure}
\hfill
\begin{subfigure}[b]{\linewidth}
\centering
\includegraphics[width=\linewidth]{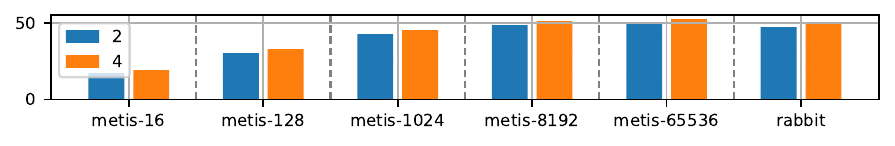}
\vspace{-6mm}
\caption{Increasing the number of layers from 2 to 4.}
\label{fig:cachemisses:cpu:layers}
\end{subfigure}
\vspace{-6mm}
\caption{Influence of increasing GNN parameters on cache miss reduction. Larger values are better.}
\label{fig:cachemisses:cpu}
\end{figure}

\textbf{CPU:} 
For CPU-based training, we observe for both \pyg{} and \dgl{} that an increasing number of hidden dimensions reduces the effectiveness of graph reordering in most cases.
For example, \rabb{} leads on average to speedups of 1.35x and 1.17x a for a hidden dimension of 16 and 256, respectively (see Figure~\ref{fig:vary-hidden-dimensions:dgl-cpu-gat-gcn}).
These results seem plausible. 
The larger the number of hidden dimensions, the larger the intermediate vertex representations.
In order to compute the representation of a vertex for the next layer, the representations of its neighbors need to be accessed.
However, the cache size is limited and the larger the representations, the fewer fit into the cache and graph reordering becomes less effective. 
To validate this hypothesis, we measure the cache miss rate for the graph reordering strategies and observe that the graph reordering strategies become indeed less effective in reducing cache misses if the number of hidden dimensions increases. 
For example, \rabb{} reduces the number of cache misses by 36\% and 31\% on \produ{} for a hidden dimension of 16 and 256, respectively (see Figure~\ref{fig:cachemisses:cpu:hidden}).

\textbf{GPU:} 
\revision{}
For GPU-based training, we make similar observations.
However, we observe one exception for GCN in \dgl{}, where graph reordering is more effective in the face of a larger hidden dimension.
Specifically, \rabb{} achieves an average speedup of 1.22x and 1.57x for a hidden dimension of 16 and 256, respectively (see Figure~\ref{fig:vary-hidden-dimensions::dgl-gpu-gcn}). 
For GAT, the trends align with those observed in CPU-based training (see Figure~\ref{fig:vary-hidden-dimensions::dgl-gpu-gat}).
This phenomenon can be attributed to GCN’s limited computational demands and its memory-bound nature, where increasing the hidden dimensions necessitates fetching more data, exacerbating memory bandwidth limitations more than computational overhead.
Therefore, reordering becomes more important with an increasing number of hidden dimensions in that scenario. 
\orginal{}
We conclude that \textbf{in most cases graph reordering becomes less effective with an increasing number of hidden dimensions, but that in GPU-based training of lightweight GNN architectures, the opposite can be the case.}

\subsubsection{Feature size}
In the following experiments, we increase the feature size from 16 to 64 and 512. 
The number of hidden dimensions is set to 16, ensuring it is not larger than the feature size. 
The number of layers is fixed at 2.

\textbf{CPU:} 
For \dgl{}, we observe in most cases, that the graph reordering strategies that lead to a speedup, become less effective in the phase of a larger feature size. 
For example, \rabb{} leads to an average speedup of 1.33x and 1.24x for a feature size of 16 and 512, respectively (see Figure~\ref{fig:vary-features-dimensions:dgl-cpu-gat-gcn}). 
For \pyg{}, we also observe that a larger feature size decreases the effectiveness of graph reordering.

Regarding the effectiveness in reducing cache misses, we observe that the larger the feature size, the less effective are the graph reordering strategies to reduce the cache misses. 
These results are plausible, as the larger the feature size, the fewer features fit into the cache and graph reordering becomes less important. For example, \rabb{} reduces the number of cache misses by 60\% and 32\% for feature sizes of 16 and 512, respectively (see Figure~\ref{fig:cachemisses:cpu:features}). 

\textbf{GPU:} 
For \dgl{}, we observe for GCN, that the effectiveness of graph reordering is decreasing with an increase in the feature size. 
For example, \rabb{}, leads to average speedups of 1.35x and 1.17x (see Figure~\ref{fig:vary-features-dimensions:dgl-gpu-gcn}) for a feature size of 16 and 512, respectively.
\blue{However, when training a GAT, we observe that the effectiveness is not much influenced by the feature size (see Figure~\ref{fig:vary-features-dimensions:dgl-gpu-gat}). 
Both observations are plausible. 
The GAT model is much more computation-heavy compared to GCN. 
Therefore, improving the locality for faster feature loading is not effective.}

We conclude that \textbf{with an increasing feature size, the effectiveness of graph reordering can decrease.} 

\subsubsection{Number of layers}
\revision{}
In the following experiments, we increase the number of layers from 2 to 3 and 4.
The feature size is set to a medium value of 64, and the number of hidden dimensions is fixed at 16, as it is typically smaller than the feature size.
\orginal{}

\textbf{CPU:} 
For CPU-based training, we observe for both \dgl{} and \pyg{} that the effectiveness of graph reordering is in most cases not much influenced by the number of layers (see Figure~\ref{fig:vary-layer-dimensions:dgl-cpu-gat-gcn}).  
The effectiveness can slightly increase or decrease, but there is \mbox{no clear trend.}

These results seem reasonable. 
In each layer, GNN computations are performed and the speedup should be similar per layer, therefore, stacking multiple layers on top of each other should not have much influence on the effectiveness of graph reordering.
We also observe that the effectiveness in reducing cache misses on \produ{} is not much dependent on the number of layers (see Figure~\ref{fig:cachemisses:cpu:layers}).

\textbf{GPU:} For GPU-based training, we make similar observations (see Figure~\ref{fig:vary-layer-dimensions:dgl-gpu-gat-gcn}). 

We conclude that \textbf{the number of layers has no clear influence on the effectiveness of graph reordering.} 

\subsection{\blue{Influence of Graph Characteristics}}
\label{sec:graph-influence-characteristics}
\blue{
In the following, we investigate how graph characteristics influence the effectiveness of graph reordering for both CPU- and GPU-based GNN training.
To enable controlled experiments, we use different graph generators to create synthetic graphs with different varying properties: 
Specifically, we vary the mean degree and the local clustering coefficient, both of which are well-known and important metrics for describing graph datasets \cite{newman2003structure}.}
\blue{The mean degree influences the density of the graph, while the clustering coefficient indicates the presence of community structures.}

\textbf{\blue{Clustering Coefficient.}} 
\blue{We use the small-world model introduced by Watts and Strogatz \cite{watts} as it can be used to generate graphs with different clustering coefficients while keeping the number of edges and vertices constant. 
We create 6 graphs with clustering coefficients of 0.09, 0.16, 0.25, 0.38, 0.53, and 0.73 (higher numbers indicate a stronger community structure). 
Each graph consists of 3 M vertices and 60 M edges.
We observe a strong correlation (Pearson correlation coefficients and Spearman rank correlation coefficients larger 0.91) between high clustering coefficients and high speedups for both \dgl{} and \pyg{} for CPU- and GPU-based training. 
This indicates that for graphs with a stronger community structure, graph reordering is more effective in speeding up GNN training.
This result seems plausible because if a graph contains a community structure, this locality can be exploited.}

\textbf{\blue{Mean Degree.}}
\blue{We use the R-MAT \cite{chakrabarti2004r} model and Barabási–Albert preferential attachment \cite{albertbarabasi} to generate graphs with different mean degrees.
For both generators, we set the number of vertices to 3 M and create 6 graphs with mean degrees ranging from 10 to 120. 
We observe that the mean degree does not correlate with the speedup for both generators, indicating that the effectiveness of graph reordering is independent of the mean degree. 
One possible reason is that while the mean degree influences the overall graph density, it does not significantly affect the locality or community structure within the graph, which are more critical for the effectiveness of reordering techniques.}

\blue{We conclude that \textbf{graph reordering is more effective for graphs with stronger community structures}.} 

\subsection{\blue{Influence of Infastructure}}
\blue{In the following, we investigate the effectiveness of graph reordering on different GPUs in \dgl{}. 
In addition, to the NVIDIA RTX 8000 GPU (Turin architecture, 48 GB memory, 672 GB/s memory bandwidth, 16.3 TFLOPS peak), we use the NVIDIA Titan X GPU (Maxwell architecture, 12 GB memory, 337 GB/s memory bandwidth, 6.7 TFLOPS peak) representing an older architecture, and an NVIDIA Orin (Ampere architecture, 48 GB memory, 205 GB/s memory bandwidth, 5.3 TFLOPS peak) representing a newer architecture.}

\blue{We observe that graph reordering is effective for all GPUs leading to average speedups of 1.13x, 1.17x, and 1.22x for Titan X, RTX 8000, and Orin, respectively. 
When employing an advanced reordering strategy such as \rabb{}, even higher average speedups of 1.19x, 1.27x, and 1.33x are achieved on the Titan X, RTX 8000, and Orin, respectively.}

\blue{These results are plausible given that GNN training is memory-bottlenecked, making data layout improvements particularly beneficial. 
One possible explanation for the varying effectiveness of graph reordering across different GPUs could be the ratio of peak FLOPS to memory bandwidth: 
We observed that for all three GPUs, the higher the ratio of FLOPS per memory bandwidth, the more beneficial the graph reordering.} 

\blue{We conclude that \textbf{graph reordering is effective across different GPUs and is especially beneficial when the ratio of computational performance to memory bandwidth is high}.}

\subsection{GPU vs. CPU-based Training}
\label{sec:ppu-vs-cpu}

\begin{figure}[t]
\centering
\begin{subfigure}[b]{\linewidth}
\centering
\includegraphics[width=\linewidth]{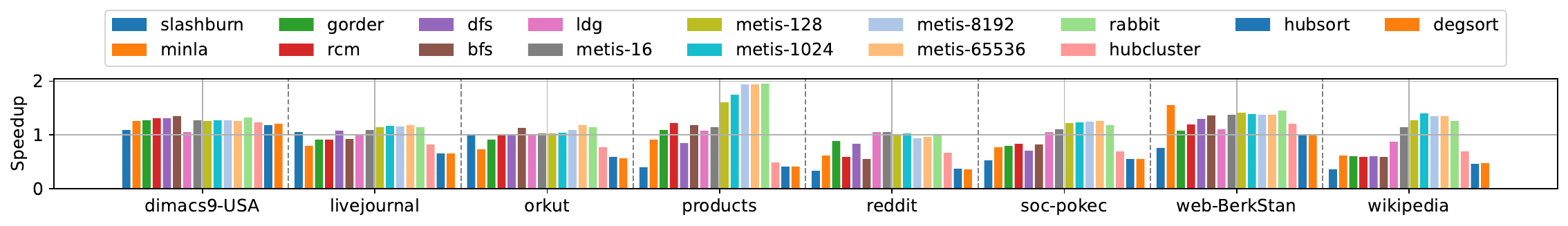}
\vspace{-6mm}
\caption{Speedups on CPU (DGL).}
\label{fig:gpu-vs-cpu:dgl-cpu}
\end{subfigure}
\hfill
\begin{subfigure}[b]{\linewidth}
\centering
\includegraphics[width=\linewidth]{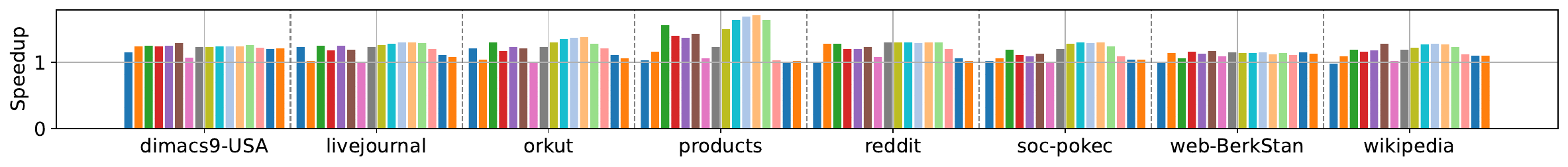}
\vspace{-6mm}
\caption{Speedups on GPU (DGL).}
\label{fig:gpu-vs-cpu:dgl-gpu}
\end{subfigure}
\hfill
\begin{subfigure}[b]{\linewidth}
\centering
\includegraphics[width=\linewidth]{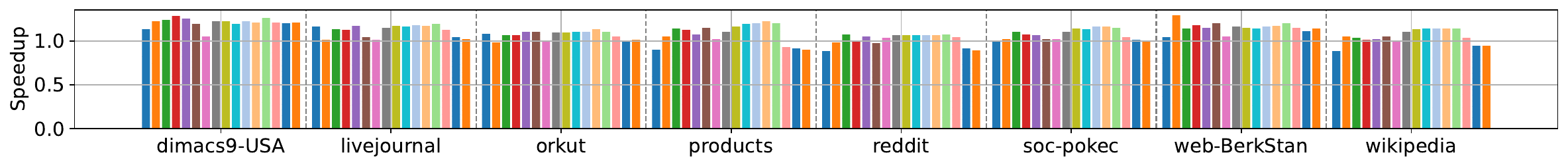}
\vspace{-6mm}
\caption{Speedups on CPU (PyG).}
\label{fig:gpu-vs-cpu:pyg-cpu}
\end{subfigure}
\hfill
\begin{subfigure}[b]{\linewidth}
\centering
\includegraphics[width=\linewidth]{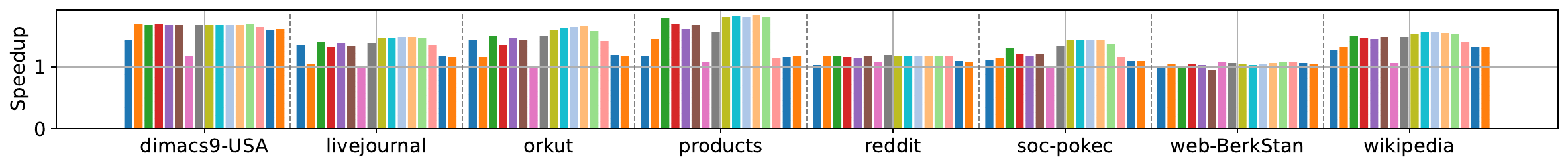}
\vspace{-6mm}
\caption{Speedups on GPU (PyG).}
\label{fig:gpu-vs-cpu:pyg-gpu}
\end{subfigure}
\vspace{-6mm}
\caption{Comparison of speedups achieved for GPU- and CPU-based training with DGL and PyG.}
%\vspace{-6mm}
\label{fig:gpu-vs-cpu}
\end{figure}

In the following, we analyze how the effectiveness of graph reordering is influenced by the infrastructure (GPU and CPU).
Due to limited GPU memory, it was not possible to run all experiments on the GPU (GPU memory is only 48 GB while the CPU server had 164 GB of memory). 
Therefore, we exclude experiments that did not run on both the CPU and GPU from the following discussion. 

\textbf{DGL:} 
We report our results for CPU- and GPU-based training in Figures~\ref{fig:gpu-vs-cpu:dgl-cpu} and \ref{fig:gpu-vs-cpu:dgl-gpu}, respectively.
We observe for the graph \dima{} that there is not much difference in terms of speedup for the different graph reordering strategies when comparing training on the CPU versus training on the GPU. 
For \dima{}, the largest difference is observed for \slas{}, leading to a speedup of 1.08x on the CPU and a higher speedup of 1.15x on the GPU. 
On the remaining graphs, we observe that some strategies (mainly \slas{}, \minla{}, \hubc{}, \hubs{}, \degs{}, \rcm{}, and \dfs{}) lead to slowdowns (speedups smaller 1x) for CPU-based training.
However, on the GPU, these strategies perform much better and lead nearly in all cases to speedups. 
On average the speedup of graph reordering for GPU-based training is 1.28 times the speedup of CPU-based training.
Interestingly, even some of the lightweight strategies can lead to speedups close to heavyweight graph reordering with \meti{} when training on the GPU.
For example, \dfs{} leads to speedups of 1.25x, 1.25x, 1.20x, 1.13x, and 1.18x on \dima{}, \live{}, \redd{}, \webb{}, and \wiki{} which is not much worse compared to \textit{metis-65536} leading to 1.24x, 1.30x, 1.30x, 1.12x, and 1.27x. 

\textbf{PyG:}
Figures~\ref{fig:gpu-vs-cpu:pyg-cpu} and \ref{fig:gpu-vs-cpu:pyg-gpu} show our results for CPU- and GPU-based training, respectively.
Similar to DGL, we observe that graph reordering is more effective for GPU-based training than for CPU-based training. 
Also similar to DGL, for GPU-based training all graph reordering strategies lead to speedups compared to CPU-based training where some led to slowdowns.
The average speedup on the GPU is 1.17 times the speedup of CPU-based training.
It can also be seen that even lightweight strategies lead to good speedups on the GPU. 
For example, \dfs{} leads to a speedup of 1.67x, 1.38x, 1.47x, 1.61x, 1.14x, 1.16x, 1.02x, and 1.44x on the graphs \dima{}, \live{}, \orku{}, \produ{}, \redd{}, \socp{}, \webb{}, and \wiki{}, respectively. 
This is close to \textit{metis-65536} with speedups of 1.67x, 1.48x, 1.66x, 1.83x, 1.17x, 1.43x, 1.06x, and 1.54x. 

\blue{The results seem plausible. 
Due to the high computational performance of GPUs, they are more memory bottlenecked compared to CPUs, making data layout improvements more beneficial.}

We conclude that \textbf{both in \dgl{} and \pyg{}, graph reordering is more effective for GPU training}. 
Further, we find that \textbf{for both systems lightweight graph reordering methods can lead to good results for GPU-based training, which was not the case for CPU-based training.} 

\begin{figure}[t]
\centering
\begin{subfigure}[b]{\linewidth}
\centering
\includegraphics[width=\linewidth]{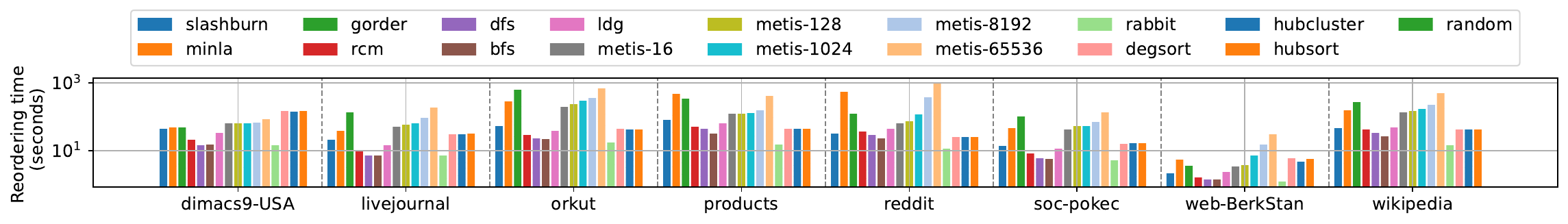}
\vspace{-6mm}
\caption{Graph reordering time (lower is better).}
\label{fig:reordering-metrics:time}
\end{subfigure}
\hfill
\begin{subfigure}[b]{\linewidth}
\centering
\includegraphics[width=\linewidth]{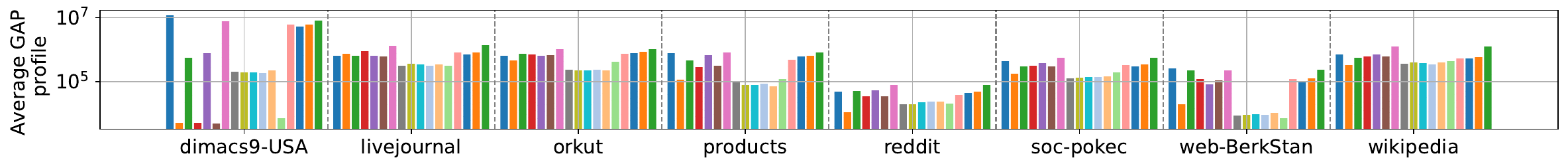}
\vspace{-6mm}
\caption{Average gap profile (lower is better).}
\label{fig:reordering-metrics:average-gap}
\end{subfigure}
\hfill
\begin{subfigure}[b]{\linewidth}
\centering
\includegraphics[width=\linewidth]{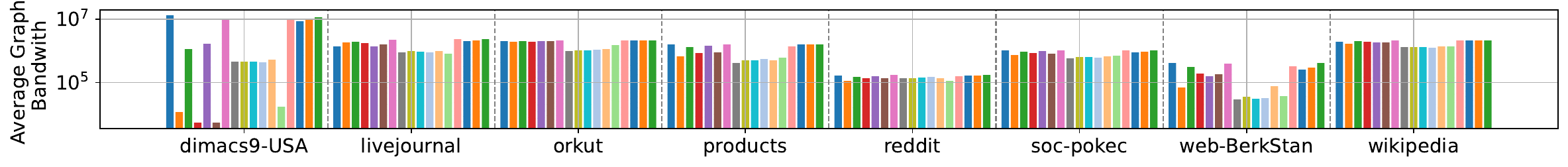}
\vspace{-6mm}
\caption{Average graph bandwidth (lower is better).}
\label{fig:reordering-metrics:average-bandwidth}
\end{subfigure}
\hfill
\begin{subfigure}[b]{\linewidth}
\centering
\includegraphics[width=\linewidth]{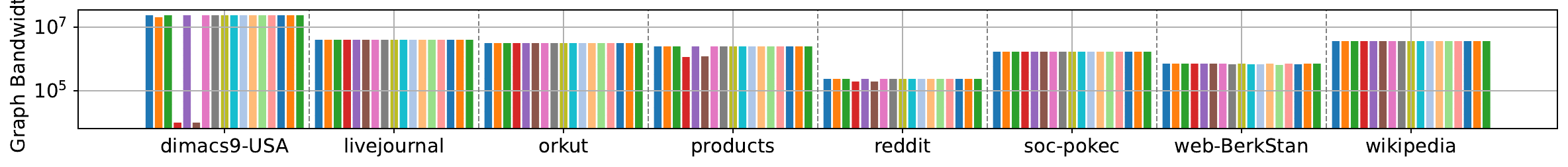}
\vspace{-6mm}
\caption{Graph bandwidth (lower is better).}
\label{fig:reordering-metrics:graph-bandwidth}
\end{subfigure}
\vspace{-7mm}
\caption{Reordering times and different reordering metrics.}
\vspace{-4mm}
\label{fig:quality-metrics}
\end{figure}

\subsection{Graph Reordering Quality Metrics}
\label{sec:quality-metrics}

In the following, we analyze how different graph reordering metrics correlate with GNN training speedup. 
We compute the graph reordering metrics \prof{}, \band{}, and \avgb{} introduced in Section~\ref{sec:background} for each combination of graph and reordering strategy and normalize it with random graph reordering. 
Then, we compute the Pearson correlation coefficient and the Spearman rank correlation coefficient between the normalized reordering metrics and training speedup.
Table~\ref{tab:correlation:metrics-speedup} reports the results.

\begin{table}[h]
\caption{Correlation between reordering metrics and training speedup measured with Pearson correlation coefficient (PC) and Spearman rank correlation coefficient (SC) and achieved speedup for selecting the best reordering strategy (Opt.), selecting the ordering strategy based on quality metrics (Qua.), and randomly selecting a reordering strategy (Ran.).}
\vspace{-4mm}
\begin{center}
\begin{tabular}{|l|c||c|c||c|c|c|}
\hline
\textbf{Metric} & \textbf{Sys.} &  \textbf{PC} & \textbf{SC} &  \textbf{Opt.} & \textbf{Qua.}& \textbf{Ran.}\\
\hline 
\hline
Avg. gap prof. & DGL & 0.58 & 0.63 & 1.39 & 1.21 & 1.10 \\ \hline
Avg. gap prof. & PyG & 0.56 & 0.64 & 1.31 & 1.24 & 1.19 \\ \hline
Avg. graph bdw. & DGL & 0.61 & 0.64 & 1.39 & 1.21 & 1.10 \\ \hline
Avg. graph bdw. & PyG & 0.58 & 0.64 & 1.31 & 1.25 & 1.19 \\ \hline
Graph bdw. & DGL & 0.11 & 0.30 & 1.39 &  1.10 & 1.10 \\ \hline
Graph bdw. & PyG & 0.13 & 0.29 & 1.31 & 1.22 & 1.19 \\
\hline
\end{tabular}
\label{tab:correlation:metrics-speedup}
\end{center}
\end{table}

We observe for both \dgl{} and \pyg{}, that the \avgb{} leads to a Pearson correlation coefficient of 0.61 and 0.58, respectively, which is slightly higher than the \prof{} for which coefficients of 0.58 and 0.56 are achieved, respectively. 
In terms of Spearman rank correlation coefficient, for \dgl{}, \avgb{} leads to a better coefficient of 0.64 compared to a coefficient of 0.63 for \prof{}. For \pyg{}, for both \prof{} and \avgb{} a coefficient of 0.64 is achieved.
This means both \prof{} and \avgb{} correlate with the speedup.
In contrast, the quality metric \band{} leads to a low Pearson correlation coefficient smaller 0.13 and a low Spearman rank correlation coefficient smaller 0.30. 
We also observe that the values for \band{} are very similar across the various graph reordering strategies, however, as reported above, different orderings lead to different speedups.  
Therefore, \band{} does not seem to be an expressive \mbox{graph reordering metric.}

\begin{figure*}[h]
\centering
\begin{subfigure}[b]{0.375\linewidth}
\centering  
\includegraphics[width=\linewidth]{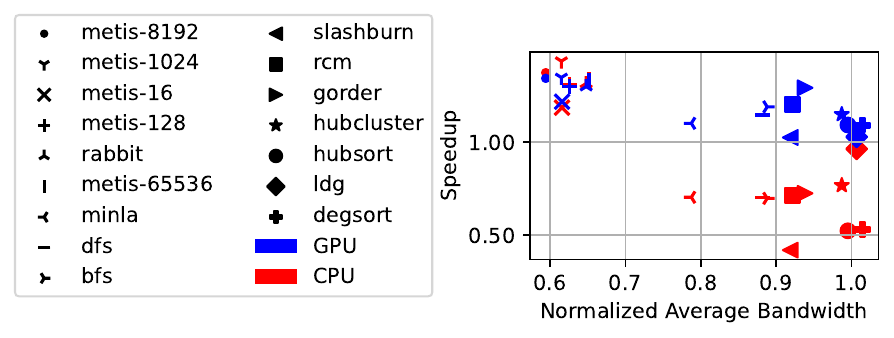}
\vspace{-6mm} 
\caption{Wiki graph (DGL).}
\label{fig:reordering-metrics:metric-vs-speedup:dgl-wiki}
\end{subfigure}
\hfill
\begin{subfigure}[b]{0.195\linewidth}
\centering
\includegraphics[width=\linewidth]{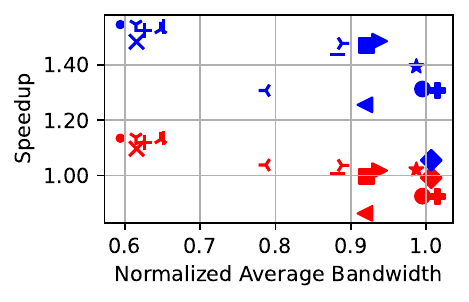}
\vspace{-6mm}
\caption{Wiki graph (PyG).}
\label{fig:reordering-metrics:metric-vs-speedup:pyg-wiki}
\end{subfigure}
\hfill
\begin{subfigure}[b]{0.195\linewidth}
\centering
\includegraphics[width=\linewidth]{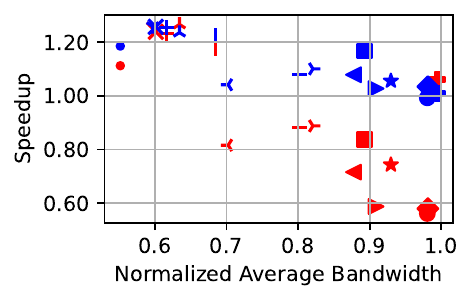}
\vspace{-6mm}
\caption{Soc-Poket graph (DGL).}
\label{fig:reordering-metrics:metric-vs-speedup:dgl-soc-pokec}
\end{subfigure}
\hfill
\begin{subfigure}[b]{0.195\linewidth}
\centering
\includegraphics[width=\linewidth]{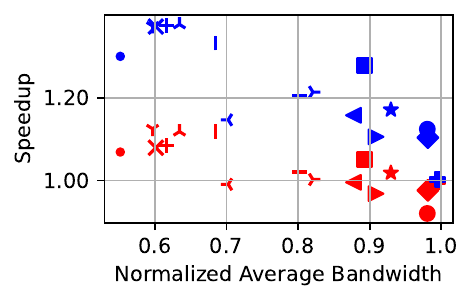}
\vspace{-6mm}
\caption{Soc-Poket graph (PyG).}
\label{fig:reordering-metrics:metric-vs-speedup:pyg-soc-pokec}
\end{subfigure}
\vspace{-3mm}
\caption{Average graph bandwidth normalized with random reordering versus speedup. The markers represent the graph reordering strategy, and the color of the markers indicates whether the training was performed on the GPU or CPU.}
\vspace{-4mm}
\label{fig:reordering-metrics:metric-vs-speedup}
\end{figure*}   

In Figure~\ref{fig:reordering-metrics:metric-vs-speedup}, we plot the \avgb{} normalized with random reordering (lower is better) against speedup for the graphs \wiki{} and \socp{} for CPU and GPU-based training using both systems \pyg{} and \dgl{} on the GCN model. 
In many cases, a small \avgb{} value also leads to larger speedups, both for GPU and CPU training, indicating that minimizing the reordering metric is beneficial (see top left in Figures~\ref{fig:reordering-metrics:metric-vs-speedup:dgl-wiki}-\ref{fig:reordering-metrics:metric-vs-speedup:pyg-soc-pokec}).
However, there are also exceptions.
For example, for the graph \wiki{}, we observe that CPU training with \gord{} leads to high speedups on \dgl{} and \pyg{} comparable to \meti{}, although the \avgb{} of \gord{} is higher (worse).
% Removed figure
For the graph \socp{}, \metii{} leads to the best (lowest) \avgb{}, but in both \dgl{} and \pyg{} other graph reorderings lead to larger speedups on CPU and GPU.  
We make similar observations for GAT. 

\revision{}
While it is plausible that in most cases a good reordering quality as measured by standard reordering metrics indicates high effectiveness, there are notable limitations. For instance, consider the metric \prof{}. Suppose we have two edges, $e_1 = (v_1, v_2)$ and $e_2 = (v_3, v_4)$, and the degree of all four vertices is 1. 
Now, let’s examine two different reorderings:
(1)	$v_1$, $v_2$, $v_3$, and $v_4$ have IDs 1K, 100K, 900K, and 990K, respectively.
(2)	$v_1$, $v_2$, $v_3$, and $v_4$ have IDs 1K, 900K, 100K, and 990K, respectively. 
This means reordering (2) only differs from reordering (1) by switching the IDs of $v_2$ and $v_3$.
The \textit{gap} for both edges would be 180K and 1780K for the first and second reordering, respectively. Thus, the second reordering appears much worse compared to the first one. 
However, in the first reordering, the connected vertices are already far apart in memory, leading to cache misses. Consequently, both reorderings might be equally suboptimal, but the metric shows a much worse score for the second one. 
This suggests that once the \textit{gap} between two vertices exceeds a certain threshold, the exact distance may no longer significantly impact performance, and the absolute value of the metric may no longer be representative.
Similar problems occur in the other reordering metrics. 
\orginal{}

We conclude that \textbf{the graph reordering metrics \avgb{} and \prof{} correlate with GNN training speedup while \band{} does not. However, there are exceptions indicating that the graph reordering metrics are not perfect, and more descriptive metrics may need \mbox{to be developed.}} 

\subsection{Graph Reordering Selection} 
In the following, we investigate if the graph reordering metrics \prof{}, \band{}, and \avgb{} can be used to select a graph reordering strategy.
For each graph, we select the graph reordering strategy that leads to the best reordering according to the quality metrics. 
This selection is independent of the GNN parameters and only depends on the graph. 
We compare this selection with the optimal selection, meaning we assume to already know for \textit{each} scenario which graph reordering strategy performs best.
The optimal selection is dependent on the GNN parameters, e.g., for a given graph not always the same graph reordering strategy is best. 
Random selection means that for each scenario a graph reordering strategy is selected randomly.  

For both \dgl{} and \pyg{}, we find that using any one of the graph reordering metrics is at least as good as a random selection~(see~Table~\ref{tab:correlation:metrics-speedup}). 
For \dgl{}, if the selection is based on \prof{}, \avgb{}, and \band{}, we achieve average speedups of 1.21x, 1.21x, and 1.1x which is between a random selection~(1.10x) and the optimal selection~(1.39x). 
For \pyg{}, if the selection is based on \prof{}, \avgb{}, and \band{}, we achieve average speedups of 1.24x, 1.25x, and 1.22x which are also between a random selection~(1.19x) and the optimal selection~(1.31x). 
\blue{For both \dgl{} and \pyg{}, \avgb{} leads to the best selections, consistent with the results from Section~\ref{sec:quality-metrics}, which also highlights the limitations of the current reordering metrics. Therefore, we anticipate that more descriptive metrics will enhance reordering selection. }

We conclude that \textbf{using the graph reordering metric \prof{} for graph reordering strategy selection outperforms a random selection, but there is still room for improvement to select the optimal strategy.}

\subsection{Graph Reordering Time Amortization}

\begin{figure*}[t]
\centering
\begin{subfigure}[b]{\linewidth}
\centering
\includegraphics[width=\linewidth]{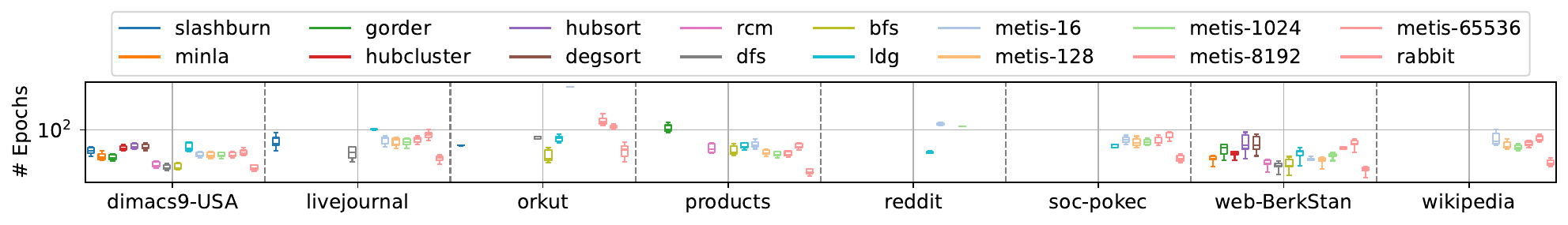}
\vspace{-6mm}
\caption{GAT on DGL (CPU).}
\label{fig:amortization:epochs:dgl-cpu-gat}
\end{subfigure}
\hfill
\begin{subfigure}[b]{\linewidth}
\centering
\includegraphics[width=\linewidth]{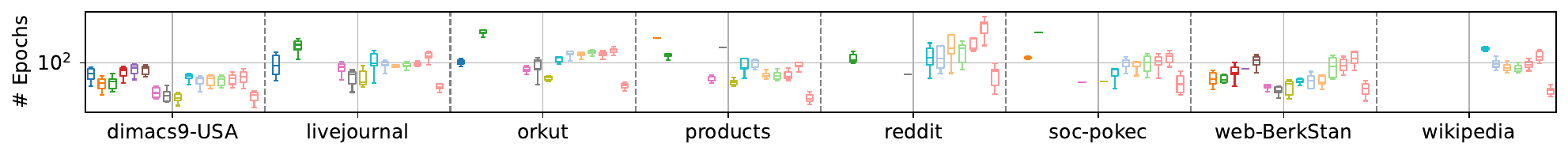}
\vspace{-6mm}
\caption{GCN on DGL (CPU).}
\label{fig:amortization:epochs:dgl-cpu-gcn}
\end{subfigure}
\hfill
\begin{subfigure}[b]{\linewidth}
\centering
\includegraphics[width=\linewidth]{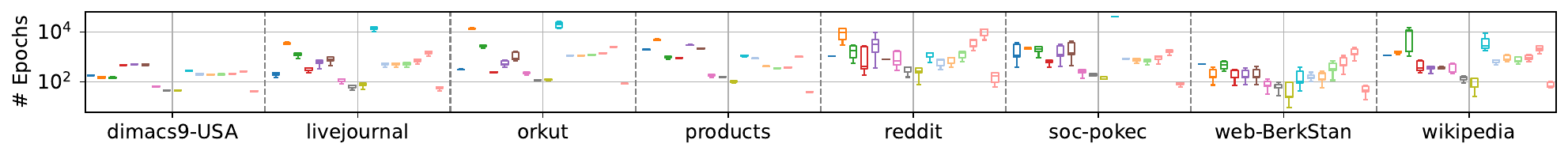}
\vspace{-6mm}
\caption{GAT on DGL (GPU).}
\label{fig:amortization:epochs:dgl-gpu-gat}
\end{subfigure}
\hfill
\begin{subfigure}[b]{\linewidth}
\centering
\includegraphics[width=\linewidth]{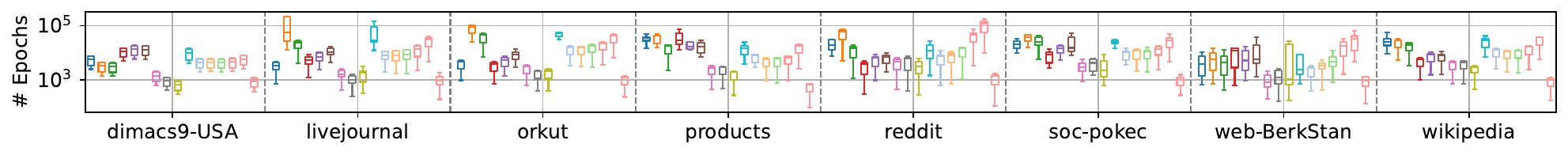}
\vspace{-6mm}
\caption{GCN on DGL (GPU).}
\label{fig:amortization:epochs:dgl-gpu-gcn}
\end{subfigure}
\hfill
\vspace{-6mm}
\caption{Number of training epochs until amortization. Lower values are better.}
\label{fig:amortization:possible}
\end{figure*}

Graph reordering is a pre-processing step that is invested to minimize graph processing time. 
In the following, we investigate after how many training epochs the graph reordering time can be amortized. 
We highlight, that training is often performed for hundreds of epochs, and often a hyper-parameter search is performed leading to even more training epochs. 
As mentioned in Section~\ref{sec:ppu-vs-cpu}, some graph reordering strategies can lead to slowdowns and therefore amortization is not possible. 
In the following, we only consider scenarios in which graph reordering leads to speedups.

\textbf{CPU:} We observe that in many cases graph reordering is amortized in far less than 100 epochs for both GAT and GCN. 
For example in \dgl{}, \metiiii{} amortizes in 4.0, 22.6, 4.5, 21.6, 3.7, and 11.1 epochs for the graphs \dima{}, \live{}, \produ{}, \socp{}, \webb{}, and \wiki{} when training a GAT and in  21.0, 77.9, 29.2, 128.4, 416.2, and 61.0 epochs when training a GCN (see Figures~\ref{fig:amortization:epochs:dgl-cpu-gat} and \ref{fig:amortization:epochs:dgl-cpu-gcn}). 
\textit{Rabbit} amortizes even faster in 0.8, 4.3, 0.4, 2.7, 0.6, and 1.6  epochs for the graphs \dima{}, \live{}, \produ{}, \socp{}, \webb{}, and \wiki{} when training a GAT and in 3.5, 12.2, 2.9, 16.7, 10.3, and 6.6 epochs when training a GCN, respectively (see Figure~\ref{fig:amortization:epochs:dgl-cpu-gcn}). 
In general training a GAT takes more time and graph reordering time can be amortized faster. 

\textbf{GPU:} We also observe that graph reordering amortizes faster for GAT than for GCN.
In general, GPU-based training is much faster than CPU-based training and graph reordering can take more time than training.
For example in \dgl{} \metiiii{} amortizes only after 188 and 4155 epochs when training a GAT and a GCN on \dima{}, respectively  (see Figures~\ref{fig:amortization:epochs:dgl-gpu-gat} and \ref{fig:amortization:epochs:dgl-gpu-gcn}).
However, it is worth noting that \rabb{} amortizes in 39.8, 55.8, 81.1, 37.0, 159.7, 80.9, 49.4, and 75.7 epochs on the graphs \dima{}, \live{}, \orku{}, \produ{}, \redd{}, \socp{}, \webb{}, and \wiki{}, respectively, when training a GAT (see Figure~\ref{fig:amortization:epochs:dgl-gpu-gat}). 
This means that even for GPU-based training graph reordering can be amortized. 
However, even more important, graph reordering can be performed on a cheap CPU server and the training can be performed on a more expensive GPU afterwards. 
Therefore, we suggest pre-processing the graphs on a CPU server to save costs when training on a GPU server. 

We conclude that \textbf{for CPU-based training graph reordering time can be amortized in many cases. For GPU-based training, the training time can not always be amortized, however, pre-processing on a CPU can still save GPU costs in the \mbox{training process.}}

\subsection{Sampling}
\revision
We also investigate the effectiveness of graph reordering when neighborhood sampling is applied.
In addition to the eight graphs in Table~\ref{tab:graphs}, we added two large-scale graphs, \twit{}~\cite{konnekt} and \pape{} from the Open Graph Benchmark~(OGB)~\cite{hu2020open}, with 1.5 billion edges and 1.6 billion edges, respectively. 
For both graphs, full graph training is infeasible on a GPU due to memory limitations, necessitating the use of neighborhood sampling.
Neighborhood sampling requires setting the fanout parameters, denoted as $(l_1, l_2,...,l_n)$, where $l_i$ represents the number of neighbors to sample in the $i^{th}$ layer. 
For our experiments, we use commonly employed fanout parameters: (25, 10) for 2-layer experiments~\cite{vldbgnnstudy,P3,graphsage}, and a fanout of (15, 10, 5) for 3-layer experiments~\cite{vldbgnnstudy,distdgl,GNNLab,salient}. 
The batch size is set to 1024.
For both \dgl{} and \pyg, we use the GCN model for CPU- and GPU-based training on a server with 512 GB main memory and a NVIDIA A40 GPU with 48 GB device memory.

\begin{figure}[t]
\centering
\begin{subfigure}[b]{\linewidth}
\centering
\includegraphics[width=\linewidth]{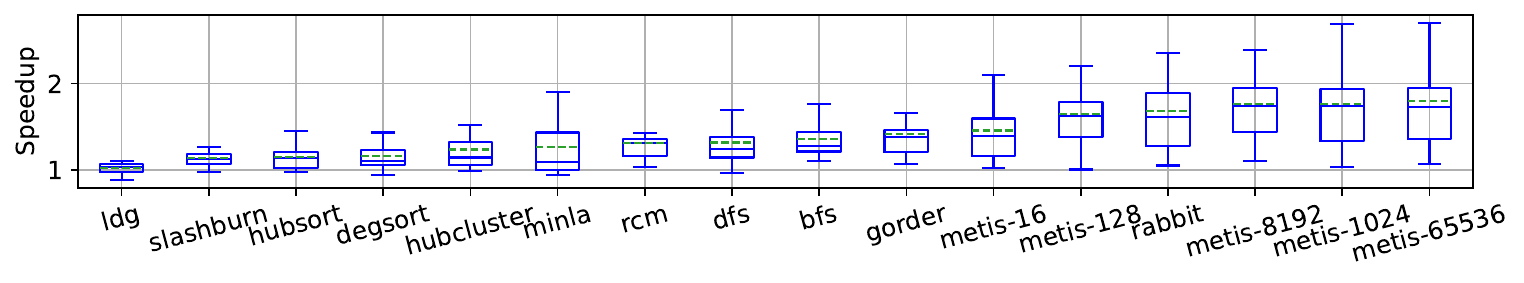}
\vspace{-6mm}
\caption{\blue{CPU-based training.}}
\label{fig:mini-batch-overview:cpu}
\end{subfigure}
\hfill
\begin{subfigure}[b]{\linewidth}
\centering
\includegraphics[width=\linewidth]{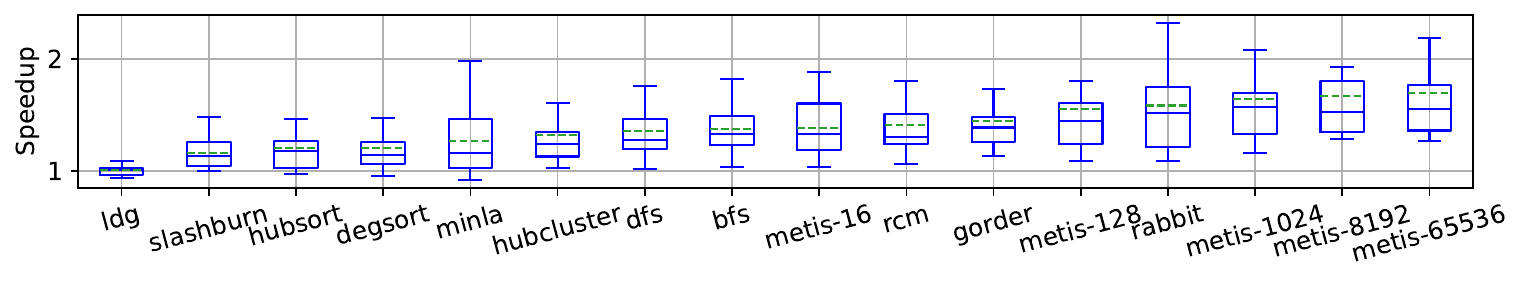}
\vspace{-6mm}
\caption{\blue{GPU-based training.}}
\label{fig:mini-batch-overview:gpu}
\end{subfigure}
\hfill
\vspace{-6mm}
\caption{\blue{Speedup distribution for sampling-based training.}}
\label{fig:mini-batch-overview}
\end{figure} 

For \pyg{}, we observe that graph reordering leads to significant speedups which are shown in Figures~\ref{fig:mini-batch-overview:cpu} and \ref{fig:mini-batch-overview:gpu} for CPU- and GPU-based training, respectively. 
When training on a CPU, effective reordering strategies such as \rabb{} and \metiiiiii{} yield average speedups of 1.68x and 1.80x, and maximum speedups of 3.35x and 3.82x, respectively.
For GPU-based training, \rabb{} and \metiiiiii{} achieve average speedups of 1.59x and 1.69x, and maximum speedups of 2.84x and 3.30x, respectively.

\begin{figure}[t]
\centering
\begin{subfigure}[b]{0.48\linewidth}
\centering
\includegraphics[width=\linewidth]{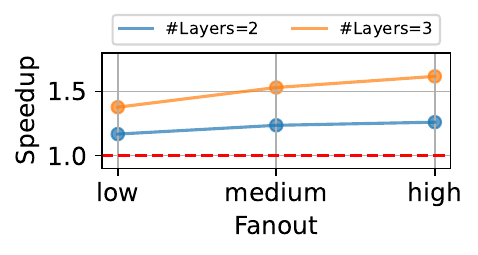}
\label{fig:sampling-fanout:cpu}
\end{subfigure}
\hfill
\begin{subfigure}[b]{0.48\linewidth}
\centering
\includegraphics[width=\linewidth]{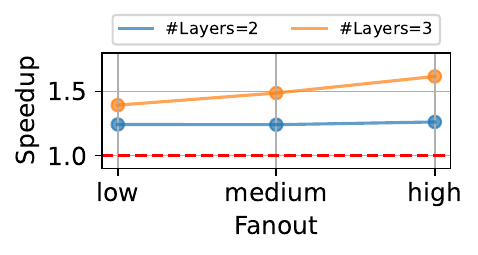}
\label{fig:sampling-fanout:gpu}
\end{subfigure}
\vspace{-8mm}
\caption{\blue{Speedup versus fanout. CPU (left) and GPU (right).}}
\vspace{-6mm}
\label{fig:sampling-fanout}
\end{figure}

In the following, we investigate how the fanout influences the effectiveness of graph reordering.
We experiment with three different fanout configurations: We use a \textit{low} fanout of (5, 5) and (5, 5, 5), a \textit{medium} fanout of (10, 10) and (10, 10, 10), and a \textit{high} fanout of (15, 15) and (15, 15, 15), for 2-layer and 3-layer GNNs, respectively. 
Figure~\ref{fig:sampling-fanout} shows the average speedup for all combinations of fanout and number of layers for CPU and GPU-based training, respectively.
We observe that graph reordering is more effective in the face of higher fanouts: 
On the CPU, the speedup increases from 1.17x to 1.26x for a 2-layer GNN and from 1.38x to 1.62x for a 3-layer GNN as the fanout increases from low to high. On the GPU, the speedup increases from 1.24x to 1.26x for a 2-layer GNN and from 1.39x to 1.62x for a 3-layer GNN.
Further, we observe that if the number of layers increases, graph reordering becomes more effective: 
For instance, on the CPU, the speedup for a low fanout increases from 1.17x to 1.38x when increasing the number of layers from two to three, and for a high fanout, it increases from 1.27x to 1.66x.
This is plausible as larger fanouts and more layers result in more sampling and feature fetching, making graph reordering more beneficial.

We further analyze which phase of the GNN training benefits from the speedup. 
GNN training with neighborhood sampling consists of two phases which are iteratively repeated: 
(1) sample a mini batch from the input graph and extract necessary features (data loading phase) and 
(2) perform the training on the sample (training phase).
For CPU-based training, we find that for a low fanout, 41\% of the speedup is achieved in the data loading phase and 59\% in the training phase, while for a high fanout, 61\% of the speedup is achieved in the data loading phase and 39\% in the training phase. For GPU-based training, these figures are 88\% and 12\% for low fanout, and 99\% and 1\% for high fanout, respectively.

For \dgl{}, we observe average speedups of 1.05x and 1.00x for CPU and GPU-based training, respectively, indicating that graph reordering is not effective to speed up GNN training in \dgl{} when neighbor sampling is applied. We attribute this phenomenon to differences in the data loading implementation in \dgl{}, which does not benefit from locality in graph ordering.

Additionally, we assess whether graph reordering time can be amortized in \pyg{}.
For CPU-based training, \rabb{} amortizes in 
0.17, 0.20, 0.33, 0.35. 0.39. 0.57, 0.84, 1.23, 3.20, and 6.1 epochs on the graphs 
\produ{}, \live{}, \socp{}, \wiki{}, \orku{}, \webb{}, \dima{}, \redd{}, \pape{}, and \twit{}, respectively. 
For GPU-based training, \rabb{} amortizes in 
0.92, 0.95, 1.21, 1.75, 1.85, 2.20, 3.67, 4.36, 6.65, and 9.87 epochs on the graphs \produ{}, \dima{}, \live{}, \webb{}, \wiki{},  \orku{}, \socp{}, \twit{}, \pape{}, and \redd{}, respectively. 
Therefore, graph reordering can be amortized for both CPU and GPU-based training, making it a practical optimization. 
Furthermore, graph reordering can be performed on a cost-effective CPU server, reducing the monetary cost of expensive GPU training time.

We conclude that \textbf{graph reordering is effective in \pyg{} when sampling is applied, but not in \dgl{}. 
Moreover, graph reordering is more effective with higher fanouts or an increased number of layers. Lastly, graph reordering time can be amortized for both CPU- and GPU-based training, highlighting its practicality as an optimization technique.}
\orginal

\section{Lessons Learned}
\label{sec:lessons-learnt}
In the following, we summarize our main findings and relate them to the research questions introduced in Section \ref{sec:experimental-methodology}.

\textbf{Graph reordering can reduce GNN training time (Q1).} 
Our experiments show that graph reordering is an effective optimization to reduce training time \blue{leading to speedups of up to 2.19x (average 1.25x) and to 2.43x (average 1.33x) for CPU- and GPU-based training, respectively, when applying \textit{Rabbit.}.}
We observe large differences between the graph reordering strategies, however, in most cases the partition-based approaches \meti{} and \rabb{} lead to the largest speedups.

\textbf{GNN-specific parameters can influence the effectiveness of graph reordering (Q2).}
For both CPU- and GPU-based training, we find that an increasing number of hidden dimensions and an increasing feature size can decrease the effectiveness of graph reordering while the number of layers has only little influence.

\textbf{Graph reordering is effective across different hardware configurations (Q3).}
We find that graph reordering can speed up GNN training on a CPU and GPU.
For both \dgl{} and \pyg{}, the achieved speedups on GPUs are larger than for CPU-based training.   
We observe that mostly the degree-based approaches can lead to a slowdown for CPU-based training, while they lead to speedups when \mbox{training on a GPU}.
\blue{Further, we find that graph reordering is effective on different GPUs, and that especially if the GPU's ratio of computation performance to memory bandwidth is high, reordering is more beneficial.}

\textbf{Graph reordering time can be amortized (Q4).}
We find that graph reordering time can be amortized in many cases when training on a CPU. 
We observed that amortization is dependent on the GNN model architecture, e.g., GAT epoch times are higher than GCN epoch times, and therefore, amortization can be achieved faster.
However, when training on a GPU, it depends on the model if graph reordering time can be amortized by faster training. 
Still, we suggest applying graph reordering when training on a GPU as the graph reordering can be performed on a cheap CPU machine, and monetary expensive training time on the GPU can be reduced.

\textbf{Graph reordering quality metrics can be used for graph reordering selection (Q5).} 
We find a correlation between graph reordering quality metrics and training speedup. 
Further, our experiments show that selecting a graph reordering strategy based on graph reordering quality metrics is much better than randomly selecting a strategy. 
However, there is room for improvement indicating that better graph reordering metrics need to be developed. 
\revision{}
As it is not straightforward to select the best reordering strategy for a given scenario, we see building an automatic, machine learning-based selector as a promising research direction. Such an approach has been successfully applied to graph partitioner selection for distributed graph processing \cite{ease}.
For practitioners, we make the following recommendation:
(1) For GPU-based training, we recommend performing the graph reordering on a cost-effective CPU machine to save monetary costs when training on the GPU. 
\meti{} should be selected if the pre-processing machine has sufficient memory, otherwise \rabb{} should be selected as it is much more light-weight in terms of memory and reordering time but still leads to good speedups. 
(2) For CPU-based training, we recommend applying \rabb{} for its quick amortization.
However, if a hyper-parameter search is performed or a more expensive model such as GAT is trained, \meti{} should be applied. 
(3) For sampling-based training in \pyg{} (CPU and GPU), we suggest applying \meti{} if enough memory is available, otherwise \rabb{} should be used.
\orginal{}

\revision{}
\textbf{Graph properties can influence the effectiveness of graph reordering (Q6).} 
We find that graph reordering is more effective for graphs with a stronger community structure, while the graph density has a lower impact.
\orginal{}

\revision{}
\textbf{Graph reordering can be effective if sampling is applied (Q7).}
In \pyg{}, graph reordering significantly speeds up GNN training, particularly by accelerating the data loading phase (sampling and feature loading). The high-quality strategy \textit{Rabbit} achieves speedups of up to 3.68x (average 1.62x) for CPU-based training and up to 3.22x (average 1.57x) for GPU-based training.
Graph reordering is more effective with higher fanouts and more layers, and reordering time can be amortized faster compared to full graph training.
\orginal{}

\section{Related Work}
\label{sec:related-work}

\textbf{Graph reordering to optimize graph analytics.}
Graph reordering is a vibrant research area~\cite{slashburn,rcm,gorder,rabbit} and different graph reordering strategies exist. 
It was shown that graph reordering can speed up graph analytics such as Breadth-first-search, Shortest Paths, or K-cores, but graph reordering time often can not be amortized by faster analytics.

In our work, we investigate graph reordering for GNN training which has unique characteristics such as high-dimensional intermediate states which are iteratively aggregated, high-dimensional feature vectors, neural network operations, and is executed for hundreds of epochs.
Further, GNN training is often performed on GPUs to accelerate computationally-extensive neural network operations.  
Our extensive evaluation of 12 different reordering strategies shows the importance of graph reordering for both CPU- and GPU-based GNN training. 
Further, we investigate graph reordering metrics and their relationship with training speedup to understand if a maximization of the metrics is beneficial to achieve training speedups. 
In addition, we investigate if quality metrics can be used for the selection of graph reordering strategies.

\textbf{Graph neural network systems.}
The recent success of GNNs led to many systems~\cite{dgl,pyg,DistGNN,distdgl,bytegnn,P3,BGL,PaGraph,Seastar,DiskGNN,10.14778/3352063.3352127,10.14778/3415478.3415482,10.14778/3538598.3538614} tailored to the specific needs of GNNs, however, it has not been investigated \blue{in which scenarios} GNN systems can benefit from graph reordering. 

In our work, we selected the two predominant systems \dgl{} and \pyg{} which are widely used and have large user communities.
For both systems, we find that graph reordering is an effective optimization to speed up the training process on CPUs and GPUs.

\blue{We anticipate that graph reordering will also be beneficial for other systems: Many systems such as DistDGL~\cite{distdgl}, GraphStorm~\cite{zheng2024graphstorm}, PaGraph~\cite{PaGraph}, P3~\cite{P3}, DistGNN~\cite{DistGNN}, BGL~\cite{BGL}, GNNLab~\cite{GNNLab}, and DiskGNN~\cite{DiskGNN} are built on top of \dgl{} or Salient~\cite{salient} which is built on top of \pyg{}, and therefore, these systems may also benefit from graph reordering. 
Further, many research papers~\cite{Seastar,PaGraph,10.1145/3437801.3441585} mention that poor data locality is a significant challenge for efficient GNN training. 
Graph reordering directly addresses this issue and therefore may be effective also in other systems.} 

\revision{}
\textbf{Optimizations to speed up GNNs.}
A recent study on GNN acceleration \cite{ijcai2022p772} identifies three graph-level optimizations: (1) graph partitioning, (2) graph sparsification, and (3) graph sampling.
Below, we relate graph reordering to these optimizations.
\orginal{}

\textbf{ (1) Graph partitioning} 
Graph partitioning~\cite{windowvertexpartitioning,restreaming, cusp, hdrf, ne, twops, hep} is a pre-processing step for distributed graph processing systems with the goal to maximize locality on the machines of a compute cluster by assigning densely connected subgraphs (partitions) to machines. 
Thereby, communication between machines is minimized as vertices that are processed together are on the same machine. 
This optimization is similar to graph reordering where vertices that are frequently accessed together are stored close by in the memory.
Different experimental studies~\cite{survey.1.vldb.2017, survey.2.vldb.2018, survey.3.vldb.2018, survey.4.sigmod.2019} investigated the effectiveness of graph partitioning for distributed graph analytics and showed that high-quality partitioning can speed up distributed graph processing. 
It was also shown that selecting a graph partitioning algorithm is challenging and that a machine learning-based selection works best~\cite{ease}. 
In a recent experimental study~\cite{merkel2023experimental}, the effectiveness of graph partitioning was investigated for distributed GNN training and showed that GNN training time can be decreased by high-quality graph partitioning.

In our work, we investigate graph reordering as a related pre-processing step that optimizes the graph data layout to speed up GNN training on a \textit{single} machine.
Further, we find that the graph partitioning-based reordering approaches \meti{} and \rabb{} can also be used for graph reordering, and are effective in reducing GNN training time by improving the data locality.
\blue{Similar to machine learning-based partitioner selection proposed in \cite{ease}, we see applying such an approach to graph reordering selection as a promising research direction.}

\revision{}
\textbf{(2) Graph Sparsification.} 
Graph sparsification reduces the graph size by dropping edges, thus decreasing memory footprint and speeding up graph processing \cite{10.1145/1374376.1374456}. 
It has been used to speedup graph processing \cite{10.1145/3295500.3356182,10.1145/3210259.3210269} and also applied to GNN training to reduce computation time and memory requirements \cite{10.1145/3459637.3482049,srinivasa2020fast}. 
Graph sparsification can also improve the prediction performance of GNNs~\cite{rong2020dropedge,9399811,pmlr-v119-zheng20d}.
 
Graph sparsification and graph reordering are orthogonal optimizations. Reordering can still be applied to enhance data locality even after sparsification.
\orginal{}

\revision{}
\textbf{(3) Graph Sampling.} 
Graph sampling reduces computational complexity and memory footprint by sampling a subset of nodes or edges for training, thus improving the efficiency and scalability of GNN training.
According to a recent study~\cite{9601152}, numerous graph sampling approaches exist that can be categorized into node-wise sampling~\cite{graphsage,10.1145/3394486.3403280}, layer-wise sampling~\cite{chen2018fastgcn,10.5555/3454287.3455296}, subgraph-based sampling~\cite{10.1145/3292500.3330925,DBLP:conf/iclr/ZengZSKP20}, and heterogenous sampling~\cite{10.1145/3292500.3330961}. 

Our study reveals that graph reordering can also accelerate sampling-based GNN training in \pyg{}, primarily by reducing data loading time.

\orginal{}

\section{Conclusions}
\label{sec:conclusions}
In our work, we investigate the effectiveness of graph reordering to reduce GNN training time. 
We find that graph reordering is an effective optimization to decrease training time for both CPU and GPU-based training.
Our experiments show that the effectiveness is influenced by GNN parameters and that the graph reordering time can be amortized. 
Further, we find that graph reordering metrics correlate with training speedup, and can be used for graph reordering selection. 
However, graph reordering metrics are not perfect predictors for speedup indicating that new graph reordering metrics may need to be developed.

%\clearpage

\bibliographystyle{ACM-Reference-Format}
\bibliography{literature}

\end{document}